\documentclass{article}

\usepackage{microtype}
\usepackage{graphicx}
\usepackage{subfigure}
\usepackage{booktabs} % for professional tables
\usepackage{dsfont}
\usepackage{hyperref}
\usepackage{caption}
\usepackage{multicol}

\usepackage[accepted]{icml2023}

\usepackage{amsmath}
\usepackage{amssymb}
\usepackage{mathtools}
\usepackage{amsthm}
\usepackage{lipsum}  

\usepackage{amsmath,amsfonts,bm}

\def\eqref#1{equation~\ref{#1}}
\def\1{\bm{1}}

\def\eps{{\epsilon}}

\def\vmu{{\bm{\mu}}}
\def\vtheta{{\bm{\theta}}}
\def\vdelta{{\bm{\delta}}}
\def\veta{{\bm{\eta}}}
\def\vmu{{\bm{\mu}}}

\def\vSigma{{\bm{\Sigma}}}

\def\vx{{\bm{x}}}

\DeclareMathAlphabet{\mathsfit}{\encodingdefault}{\sfdefault}{m}{sl}
\SetMathAlphabet{\mathsfit}{bold}{\encodingdefault}{\sfdefault}{bx}{n}

\DeclareMathOperator*{\argmin}{arg\,min}

\usepackage[capitalize,noabbrev]{cleveref}

\theoremstyle{plain}

\theoremstyle{definition}

\theoremstyle{remark}

\usepackage[textsize=tiny]{todonotes}
\renewcommand{\algorithmiccomment}[1]{#1} 

\icmltitlerunning{Adversarial robustness of amortized Bayesian inference}

\begin{document}

\twocolumn[
\icmltitle{Adversarial robustness of amortized Bayesian inference}

% It is OKAY to include author information, even for blind
% submissions: the style file will automatically remove it for you
% unless you've provided the [accepted] option to the icml2023
% package.

% List of affiliations: The first argument should be a (short)
% identifier you will use later to specify author affiliations
% Academic affiliations should list Department, University, City, Region, Country
% Industry affiliations should list Company, City, Region, Country

% You can specify symbols, otherwise they are numbered in order.
% Ideally, you should not use this facility. Affiliations will be numbered
% in order of appearance and this is the preferred way.
\icmlsetsymbol{equal}{*}

\begin{icmlauthorlist}
\icmlauthor{Manuel Gloeckler}{tue}
\icmlauthor{Michael Deistler}{tue}
\icmlauthor{Jakob H. Macke}{tue,mpi}
\end{icmlauthorlist}

\icmlaffiliation{tue}{Machine Learning in Science, University of Tübingen and Tübingen AI Center, Tübingen, Germany}
\icmlaffiliation{mpi}{Max Planck Institute for Intelligent Systems, Department Empirical Inference, Tübingen, Germany}

% TODO others

\icmlcorrespondingauthor{Manuel Gloeckler}{manuel.gloeckler@uni-tuebingen.de}

% You may provide any keywords that you
% find helpful for describing your paper; these are used to populate
% the "keywords" metadata in the PDF but will not be shown in the document
\icmlkeywords{Machine Learning, ICML}

\vskip 0.3in
]

% this must go after the closing bracket ] following \twocolumn[ ...

% This command actually creates the footnote in the first column
% listing the affiliations and the copyright notice.
% The command takes one argument, which is text to display at the start of the footnote.
% The \icmlEqualContribution command is standard text for equal contribution.
% Remove it (just {}) if you do not need this facility.

%\printAffiliationsAndNotice{}  % leave blank if no need to mention equal contribution
\printAffiliationsAndNotice{} % otherwise use the standard text.

\begin{abstract}

% In recent years, an increasing amount of Bayesian inference problems are solved by neural networks in an amortized fashion. 
% Especially in the field of simulation-based inference advances in neural density estimation made flexible approximate Bayesian inference available for complex stochastic simulation models.
% %
% Yet, it is well known that neural networks are generally vulnerable to adversarial examples i.e. small perturbations designed to fool the network.
% %
% In this study, we examine the behavior of amortized Bayesian inference algorithms when faced with adversarial attacks.
% %
% We find that even slight perturbations can significantly distort the approximate posterior.
% %
% As a result, we investigate the use of adversarial defense strategies that can effectively protect against adversarial examples, though at the cost of potentially producing more conservative estimates on clean data.

 % Jakob's version 26th of Jan.
Bayesian inference usually requires running potentially costly inference procedures separately for every new observation. In contrast, the idea of \emph{amortized} Bayesian inference is to initially invest computational cost in training an inference network on simulated data, which can subsequently be used to rapidly perform inference (i.e., to return estimates of posterior distributions) for new observations. %This can, for example, be achieved by training a conditional density estimation network on simulated data. 
This approach has been applied to many real-world models in the sciences and engineering, but it is unclear how robust the approach is to adversarial perturbations in the observed data. %Neural networks are well-known to be vulnerable to adversarial attacks, i.e., to perturbations that are targeted to change the neural network prediction. 
Here, we study the adversarial robustness of amortized Bayesian inference, focusing on simulation-based estimation of multi-dimensional posterior distributions. 
We show that almost unrecognizable, targeted perturbations of the observations can lead to drastic changes in the predicted posterior and highly unrealistic posterior predictive samples, across several benchmark tasks and a real-world example from neuroscience.
We propose a computationally efficient regularization scheme based on penalizing the Fisher information of the conditional density estimator, and show how it improves the adversarial robustness of amortized Bayesian inference.

\end{abstract}

% Notes

% -- need to improve visualization of results, in particular figure 1

% -- need to be clear what we do (only differentiable summary statistics, gradient-based attacks) and discuss limitations/scope

% -- lets only do L2

% -- figures need a lot of work (alignment etc)

% -- need to discuss Max's paper

% -- get rid of URLS in citations

% -- possibly move some examples to appendix 

\section{Introduction}
\label{submission}
% A principled way to identify unknown parameters that match empirical observations is Bayesian inference. Yet, being Bayesian often comes with an increased computational burden.  The most common methods are Variational Inference (VI) and Markov chain Monte Carlo (MCMC), which only require a likelihood function. Yet if inference must be performed for many observations, these methods may also become computationally expensive. Thus several methods were developed to \textit{amortize} the cost of Bayesian inference. ...

% TWO PARTS THAT USE AI:
% - PART I: VAE  -> rKL ? (MAYBE...)
% - PART II: SBI -> fKL

Bayesian inference is a commonly used approach for identifying model parameters that are compatible with empirical observations and prior knowledge. Classical Bayesian inference methods such as Markov-chain Monte Carlo (MCMC) can be computationally expensive at test-time, as they rely on repeated evaluations of the likelihood function and, therefore, require a new set of likelihood evaluations for each observation.
In contrast, the idea of \textit{amortized} Bayesian inference is to approximate the mapping from observation to posterior distribution by a conditional density estimator, often parameterized as a neural network. 
Once this density estimation network has been trained, inference on a particular observation can be performed very efficiently, requiring only a single forward-pass through the network. 
This \textit{amortization} can be achieved by training conditional density estimators on simulated data and framing Bayesian inference as a prediction problem: For \textit{any} observation, the neural network is trained to predict either the posterior directly \citep{papamakarios_fast_2016, greenberg_automatic_2019, goncalves_training_2020, radev_bayesflow_2020} or a quantity that allows to infer the posterior without further simulations \citep{papamakarios_sequential_2019, hermans_likelihood-free_2020}. This approach has several advantages over MCMC methods: It can be used to perform `simulation-based inference', i.e., applied to models which are only implicitly given as simulators (models which allow to sample the likelihood but not to evaluate it), it does not require the model to be differentiable (as compared to, e.g., Hamiltonian Monte Carlo), and it allows application in high-throughput scenarios \citep{dax2021real, von2022mental,  boelts_flexible_2022, arnst_hybrid_2022}. 
%In recent years, several approaches that utilize neural density estimation for Bayesian inference have been developed, such as 
%\citep{papamakarios_fast_2018, lueckmann_flexible_2017, greenberg_automatic_2019, radev_bayesflow_2020, geffner_score_2022, sharrock_sequential_2022}. These methods, known as simulation-based inference (SBI), are particularly useful for situations where the likelihood of the generative model is intractable. They also offer the advantage of amortized per-sample inference cost, thanks to the use of a flexible conditional density estimator $q_\phi(\vtheta | \vx)$. This estimator is trained to approximate the posterior distribution $p(\vtheta | \vx)$ with reasonable accuracy using simulations of a complex 'simulator' for all data points $\vx \sim p(\vx)$ \cite{lueckmann_benchmarking_2021}. Such methods, which leverage the expressiveness and flexibility of neural networks, have gained widespread adoption in various scientific fields \cite{dax_amortized_nodate, goncalves_training_2020, deistler2022energy, boelts_flexible_2022, arnst_hybrid_2022}.

However, these benefits come at a cost: the posterior predicted by the neural network will not be exact \citep{lueckmann_benchmarking_2021}, can be overconfident \citep{hermans_trust_2022}, and can be sensitive to misspecified models \citep{cannon_investigating_2022, schmitt_detecting_2022}. Here, we study another possible limitation of neural network-based amortized Bayesian inference: It is well known that neural networks can be susceptible to adversarial attacks, i.e., tiny but targeted perturbations to the inputs can lead to vastly different outputs \citep{szegedy_intriguing_2014}.
For amortized Bayesian inference, this would indicate that even minor perturbations in the observed data could lead to entirely different posterior estimates.

Adversarial attacks have become a common technique to evaluate the robustness of ML algorithms.
%, even when a real adversary is unlikely, as is often the case in scientific applications.
Attacks can be used to assess performance in the presence of small worst-case perturbations, offering valuable insights into how models perform when faced with model misspecification. Furthermore, amortized inference is increasingly used in real-world safety-critical applications such as, e.g., robotics \citep{ramos2019bayessim} or applications accessible to the general public \citep{moon2023amortized, shen-etal-2023-reliable}. In science and engineering, users are usually domain experts, but they are often not machine learning experts and, hence, must be aware of the limitations and brittleness of any such methods.

Here, we investigate the impact of adversarial attacks on amortized inference, focusing on a particular method for amortized Bayesian inference, namely Neural Posterior Estimation (NPE, \citealt{cranmer2020frontier}). While adversarial attacks have been extensively studied in the context of classification \citep{rauber2017foolbox, croce_robustbench_2021, li_review_2022}, we present an approach and benchmark problems for evaluating the adversarial robustness of neural networks approximating multi-dimensional Bayesian posterior distributions. Using this approach, we demonstrate that NPE can be highly vulnerable to adversarial attacks. Finally, we develop a computationally efficient method for improving the adversarial robustness of NPE, and demonstrate its utility on a real-world example from neuroscience.

\begin{figure}
    \centering
    \includegraphics[width=.5\textwidth]{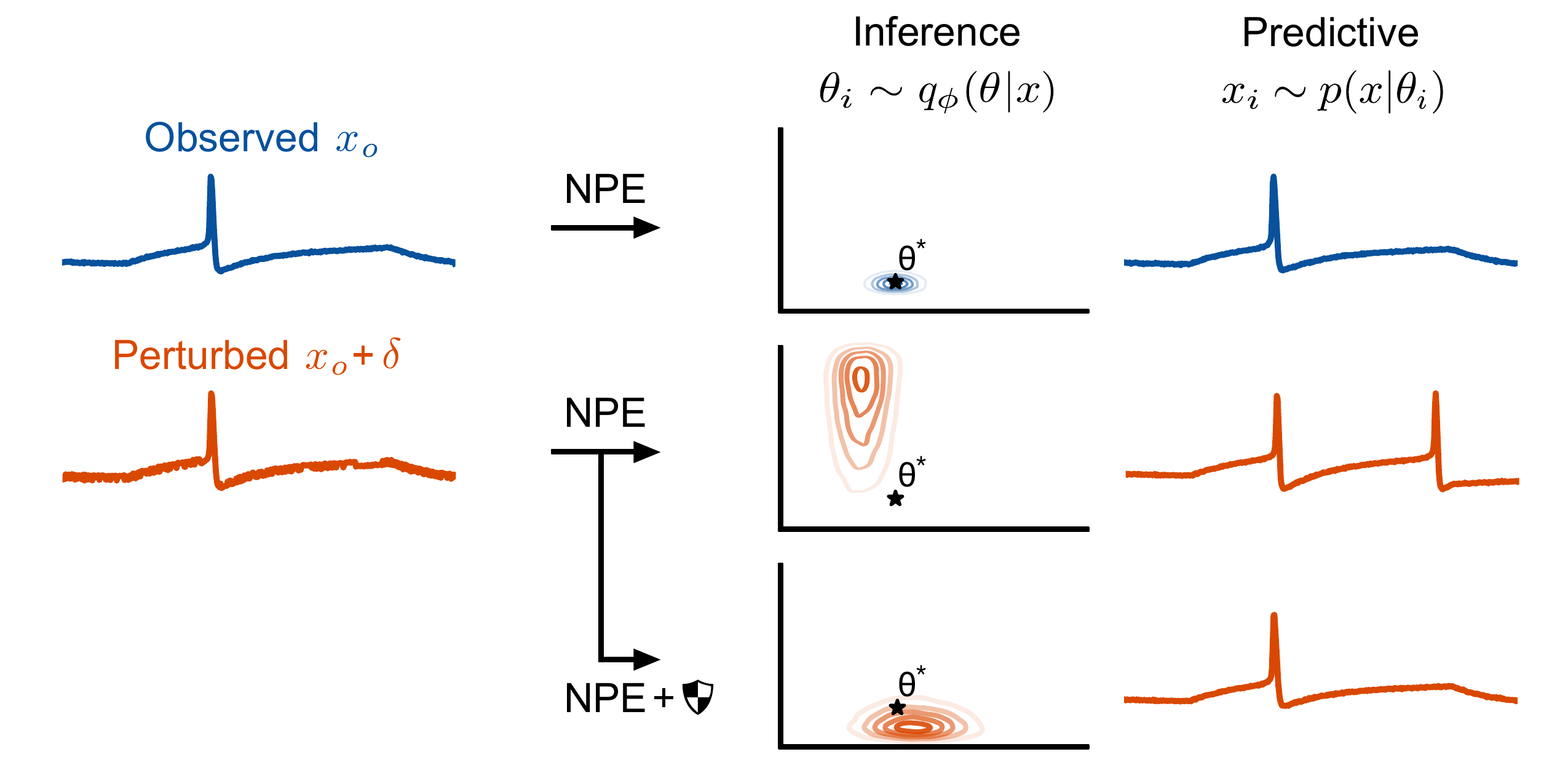}
    \caption{{\bf Adversarial attack on amortized inference.} A minor perturbation of the observed data (left column, here: a voltage recording) creates a remarkably different estimate of the posterior over parameters (middle column, here: over parameters of a biophysical neuron model). Predictive samples from the adversarial posterior estimate are very different \emph{both} from the observed and perturbed data (right column)--in this case, they exhibit two ``spikes'', while the original data only has a single one---showing that the attack leads to a break-down of the posterior \emph{estimate} of the inference network (rather than a change in the true posterior). Our defense strategy leads to a more reliable posterior estimate (bottom row) with realistic predictive samples.}
    \label{fig:fig1}
\end{figure}

Our overall approach is the following (Fig.~\ref{fig:fig1}):
Given an observation $\vx_o$, we consider an adversarial perturbation (Sec.~\ref{sec:adversarial_attacks}). As we will show, even barely visible adversarial perturbations can strongly change estimated posterior distributions, and lead to predictive samples which strongly deviate from the original observation. We suggest and implement a defense strategy (Sec.~\ref{sec:adversarial_defense}), and will show that it reduces the impact on the posterior estimate, in particular, such that it still contains the ground truth parameters.

%\section{Related work}

%but this requires likelihood evaluations after having trained the neural network. 

\section{Background and Notation}

\subsection{Amortized Bayesian inference}

In this work, we consider a fixed generative model that defines a relationship between $\vx$ and unknown parameters $\vtheta$, given by $p(\vtheta, \vx) = p(\vx|\vtheta)p(\vtheta)$. By Bayes theorem, there exists a function $f: \mathcal{X} \rightarrow \mathcal{P}(\vtheta)$ which maps data onto the posterior distribution $f(\vx_o) = p(\vtheta|\vx_o)$. As opposed to computing the posterior distribution for every observation, amortized Bayesian inference targets to learn the mapping $f$ directly, thereby amortizing the cost of inference.

One method to perform amortized Bayesian inference is Neural Posterior Estimation (NPE). NPE first draws samples from the joint distribution $p(\vtheta, \vx)$ and then trains a conditional density estimator $q_{\phi}(\vtheta | \vx)$ with learnable parameters $\phi$ to approximate the posterior distribution:
\begin{align*}
\mathcal{L}(\phi) = \mathbb{E}_{p(\vtheta, \vx)} \left[ -\log q_{\phi}(\vtheta | \vx) \right] \approx \frac{1}{N} \sum_{i=1}^N -\log q_{\phi}(\vtheta_i | \vx_i)
\end{align*}
If the conditional density estimator is sufficiently expressive, then this is minimized if and only if $q_{\phi}(\vtheta | \vx) = p(\vtheta | \vx)$ for all $\vx$ in the support of $p(\vx)$ \citep{papamakarios_fast_2016}.

\subsection{Adversarial attacks and defenses}
\label{sec:background_attack_defense}

\citet{szegedy_intriguing_2014} first proposed the concept of adversarial examples to fool neural networks. Adversarial examples are typically defined as solutions to an optimization problem \citep{szegedy_intriguing_2014, goodfellow_explaining_2015}
\begin{equation*}
    \tilde{\vx} = \arg\max_{||\tilde{\vx} - \vx||_\mathcal{X} \leq \epsilon} \Delta(f(\tilde{\vx}), f(\vx)),
\end{equation*} 
where $\Delta$ specifies a distance between the predictions of the neural network.

Many defenses against adversarial examples have been proposed. %Adversarial training \citep{madry_towards_2019} incorporates adversarial examples into the training process and is currently considered a state-of-the-art defense for classification tasks \citep{croce_robustbench_2021, rebuffi_fixing_2021}. 
We build upon a popular defense called TRADES \citep{zhang_theoretically_2019}-- when translated to inference tasks, TRADES can be interpreted as regularizing the neural network loss with the Kullback-Leibler divergence between the clean data and an adversarially perturbed data point:
\begin{equation*}
\begin{split}
\mathcal{L}(\phi)= & \mathbb{E}_{p(\tilde{\vx},\vx,\vtheta)} [ - \log q_\phi (\vtheta|\vx ) + \\ 
& \beta D_{KL}(q_\phi(\vtheta|\vx)||q_\phi(\vtheta|\tilde{\vx})) ]
\end{split}
\end{equation*}
Here, $\tilde{\vx}$ is obtained by generating an adversarial example \emph{during training}. This regularization requires generating an adversarial example for every datapoint and epoch, which requires running several gradient descent steps for every datapoint $\vx$-- this would be exceedingly computationally costly for our inference tasks, but we will present methods for overcoming this limitation.
To simplify notation, we abbreviate the posterior estimate given clean data as $q := q_{\phi}(\vtheta | \vx)$ and given perturbed data as $\tilde{q} := q_{\phi}(\vtheta | \tilde{\vx})$. %where necessary.

\section{Methods}

\subsection{Adversarial attacks on amortized inference}
\label{sec:adversarial_attacks}

%Adversarial perturbations are small, often imperceptible changes made to input data that cause a machine learning model to make an incorrect prediction.
Adversarial perturbations are typically studied in classification tasks, in which the perturbation makes the neural network predict a wrong class. For amortized Bayesian inference, however, the output of the neural network is a continuous probability distribution (the estimate of the posterior). We therefore define the target of the adversarial perturbation to maximize the divergence between the estimated posterior given the `clean' vs. the adversarially perturbed data, i.e., $D_{KL}(q(\vtheta| \vx)|| q(\vtheta| \vx + \vdelta))$ \citep{gondim-ribeiro_adversarial_2018, willetts_improving_2021, dax_neural_2022,dang-nhu_adversarial_2020}.

We here focus on the Kullback-Leibler divergence\footnote{We focus on  $D_{KL}(q||\tilde{q})$ to generate and evaluate attacks, but we discuss and evaluate the effect of a different adversarial objective in %Appendix 
Sec.~\ref{sec:attack_additional_results}.}, but any divergence or pseudo-divergence (e.g. a distance function on moments of the posterior) would be possible \citep{gondim-ribeiro_adversarial_2018, willetts_improving_2021, dax_neural_2022, dang-nhu_adversarial_2020}. 
%To understand the impact of $D$, we consider both the forward $D_{KL}(\tilde{q}||q)$ and backward divergences $D_{KL}(q||\tilde{q})$.
An attack is thus defined by the constrained optimization problem
\begin{equation*}
     \vdelta^* = \arg \max_{\vdelta} D_{KL} \left(q_\phi(\vtheta|\vx) || q_\phi(\vtheta |\vx + \vdelta)\right) \text{ s.t. } ||\vdelta|| \leq \epsilon.
\end{equation*}
To solve it, we use projected gradient descent (PGD) as an attacking scheme \citep{madry_towards_2019}, following work on adversarial robustness for classification. We estimate the divergence between distributions parameterized by conditional normalizing flows using Monte Carlo sampling. We use the reparameterization trick \citep{kingma_auto-encoding_2022} to estimate gradients (details in %Appendix 
\ref{sec:appendix_training_procedure}).

 We note that small perturbations to the observed data are expected to change the \emph{true} posterior distribution. A sufficiently small perturbation will, in general, only cause a minor change in the posterior distribution \citep{latz_well-posedness_2020}. Furthermore, posterior predictive samples should match the perturbed observation \citep{berger_overview_1994, sprungk_local_2020}. In contrast, we will demonstrate that the estimated posterior will change strongly after minor changes to the data, and that predictive samples of the posterior estimate do not match the perturbed observation, implying that the attack indeed breaks the amortized posterior \emph{estimate}.% by the neural network.% does not necessarily inherit this property, we expect it to change more than it should. 

\subsection{An adversarial defense for amortized inference}
\label{sec:adversarial_defense}
How do we modify NPE to be robust against such attacks? As described in Sec.~\ref{sec:background_attack_defense}, many adversarial defenses (e.g., TRADES) rely on generating adversarial examples during training, which can be computationally costly. In particular, for expressive conditional density estimators such as normalizing flows, generating an adversarial attack requires several Monte Carlo (MC) samples at every gradient step, thus rendering this approach exceedingly costly. Here, we propose a computationally efficient method based on a moving average estimate of the trace of the Fisher information matrix.

\paragraph{Regularizing by the Fisher information matrix}

To avoid having to generate adversarial examples during training, we exploit the fact that adversarial perturbations tend to be small and apply a second-order Taylor approximation to the KL-divergence (as has been done in previous work, \citealt{zhao_adversarial_2019, shen_defending_2019, miyato_distributional_2016}). This results in a quadratic expression \citep{blyth_local_1994},
$$ D_{KL}(q_\phi(\vtheta|\vx)||q_\phi(\vtheta|\vx + \vdelta)) \approx \frac{1}{2} \vdelta^T \mathcal{I}_\vx \vdelta, $$
where $\mathcal{I}_\vx$ is the Fisher information matrix (FIM) with respect to $\vx$, which is given by
\begin{equation*}
    \mathcal{I}_\vx = \mathbb{E}_{q_\phi(\vtheta|\vx)} \left[ \nabla_\vx \log q_\phi (\vtheta|\vx) (\nabla_\vx \log q_\phi(\vtheta | \vx))^T \right].
\end{equation*} 
This suggests that the neural network is most brittle along the eigenvector of the FIM with the largest eigenvalue (in particular, for a linear Gaussian model, the optimal attack on $D_{KL}$ corresponds \emph{exactly} to the largest eigenvalue of the FIM, Sec.~\ref{appendix:gaussian_linear}).). To improve robustness along this direction, one can regularize with the largest eigenvalue of the FIM $\lambda_{\text{max}}$ \citep{zhao_adversarial_2019, shen_defending_2019, miyato_distributional_2016}:
\begin{align*}
\mathcal{L}(\phi)= \mathbb{E}_{p(\vx,\vtheta)} \left[ - \log q_\phi (\vtheta|\vx ) + \beta \lambda_{\text{max}} \right].
\end{align*}
While this approach overcomes the need to generate adversarial examples during training, computing the largest eigenvalue of the FIM can still be costly: First, it requires estimating an expectation over $q_\phi(\vtheta|\vx)$ to obtain the FIM and, second, computing the largest eigenvalue of a potentially large matrix. Below, we address these challenges.

\paragraph{Reducing the number of MC samples with moving averages}

For expressive density estimators such as normalizing flows, the expectation over $q_\phi(\vtheta|\vx)$ cannot be computed analytically, and has to be estimated with MC sampling:
\begin{equation*}
    \hat{\mathcal{I}}_\vx = \frac{1}{N} \sum_i \left[ \nabla_\vx \log q_\phi (\vtheta_i|\vx) (\nabla_\vx \log q_\phi(\vtheta_i | \vx))^T \right] 
\end{equation*}
To reduce the number of samples required, we exploit that consecutive training iterations result in small changes of the neural network, and use an exponential moving average estimator for the FIM, i.e., 
%\begin{equation*}
  $  \hat{\mathcal{I}}_\vx^{(t)} = \gamma \hat{\mathcal{I}}_\vx + (1-\gamma) \hat{\mathcal{I}}_\vx^{(t-1)}$,
%\end{equation*}
where the superscript $(t)$ indicates the training iteration.

\paragraph{Using the trace of the Fisher information matrix as regularizer}
Such an exponential moving average estimator decreases the number of required MC samples, but it would require storing the FIM for each $\vx$ and computing the FIM's largest eigenvalue at every iteration. Computing the largest eigenvalue scales cubically with the number of dimensions of $\vx$ (but could be scaled with power-iterations, \citealt{miyato_distributional_2016}) and obtaining the largest eigenvalue of a random matrix (such as the MC-estimated FIM) requires many MC samples \citep{hayashi_bias_2018,hayou_overestimation_2017}. To overcome these limitations, we regularize instead with the trace of the FIM, which is an upper bound to the largest eigenvalue.

Unlike the largest eigenvalue, the trace of the FIM can be computed from MC samples quickly and without explicitly computing the FIM. Using the trace of the FIM simplifies the moving average estimator to
%\begin{equation*}
    $$\mbox{tr}(\hat{\mathcal{I}}_\vx^{(t)}) = \gamma \mbox{tr}(\hat{\mathcal{I}}_\vx) + (1-\gamma) \mbox{tr}(\hat{\mathcal{I}}_\vx^{(t-1)}).$$
%\end{equation*} 

% To overcome this, we use that the trace is a linear operator and, thus, $r := \mathbb{E}_{p(\vx)}{[tr(\hat{\mathcal{I}}_\vx)]} = tr(\mathbb{E}_{p(\vx)}{[\hat{\mathcal{I}}_\vx]})$. With this, can update the regularizer while only storing the \emph{average} trace of the FIM $r^{(t)}$:
% \begin{equation}
%     r^{(t)} = \gamma \mathbb{E}_{p(\vx)} \left[ {tr(\hat{\mathcal{I}}_\vx)} \right] + (1-\gamma) r^{(t-1)}
% \end{equation}

To avoid maintaining the computation graph for every $\vx$ and $(t)$, we store the \emph{average} gradient with respect to the neural network parameters instead of storing $\mbox{tr}(\hat{\mathcal{I}}_\vx^{(t)})$ directly,
\begin{equation*}
    g^{(t)} := \nabla_{\phi} \mathbb{E}_{p(\vx)} \left[ \mbox{tr}(\hat{\mathcal{I}}_\vx^{(t)}) \right].
\end{equation*}

\begin{algorithm}[t]
 \begin{algorithmic}
    \STATE{\textbf{Inputs:}} conditional density estimator $q_{\phi}(\vtheta | \vx)$ with learnable parameters $\phi$, batch size $B$, number of training steps $T$, learning rate $\alpha$, regularization strength $\beta$, regularization momentum $\gamma$, number of Monte Carlo samples $N$
    \STATE{\textbf{Initialize:}}  $g^{(0)} = 0$
    \FOR{$t=1$ {\bfseries to} $T$}
        \FOR{$b=1$ {\bfseries to} $B$}
        \STATE $\mathcal{L}(\phi) = - \frac{1}{B} \log q_\phi (\vtheta_b | \vx_b)$  \COMMENT{// NPE loss}\\
        \STATE $\vtheta_{1b}, \dots, \vtheta_{Nb} \sim q_{\phi}(\vtheta | \vx_b)$  \COMMENT{// Monte Carlo} \\
        \ENDFOR
    \STATE $r = \frac{1}{B} \sum_b \frac{1}{N} \sum_i \sum_d \left[\nabla_\vx \log q_\phi (\vtheta_{ib}|\vx_b)\right]^2_d$ \\ \qquad \qquad \qquad  \qquad \qquad \qquad \qquad 
 \  \COMMENT{// FIM Trace} \\
    \STATE $g^{(t)} = \gamma \nabla_\phi r + (1-\gamma) g^{(t-1)}$ \COMMENT{// moving average} \\
    \STATE $\phi^t = \phi^{t-1} - \alpha(\text{ADAM}(\nabla_\phi \mathcal{L}(\phi) + \beta g^{(t)}))$ 
    \ENDFOR
\end{algorithmic}
\caption{FIM-regularized NPE}
\label{alg:alg1}
\end{algorithm}

\paragraph{Summary and illustration}
Our adversarial defense is summarized in Algorithm \ref{alg:alg1}. At every iteration, the method computes the Monte Carlo average of the trace of the Fisher information, updates the moving average of this quantity, and uses it as a regularizer to the negative log-likelihood loss. Despite our approximations, our method performs similarly to regularizers based on the largest eigenvalue or trace of the exact FIM (comparison with a Gaussian density estimator on the VAE task in Sec.~\ref{sec:FIM}). 
 Finally, we note that using the FIM-regularizer systematically changes the posterior estimate even with infinite training data and, therefore, leads to a trade-off between accuracy on clean data and robustness to perturbations (Sec.~\ref{sec:appendix_tradeoff}). For a generalized linear Gaussian density estimator, the bias induced by FIM-regularization can be calculated exactly (details in Sec.~\ref{appendix:defense_analytic}).
%~ Sec.~\ref{sec:appendix_tradeoff}).

\begin{figure}[t]
    \centering
    \includegraphics[width=.5\textwidth]{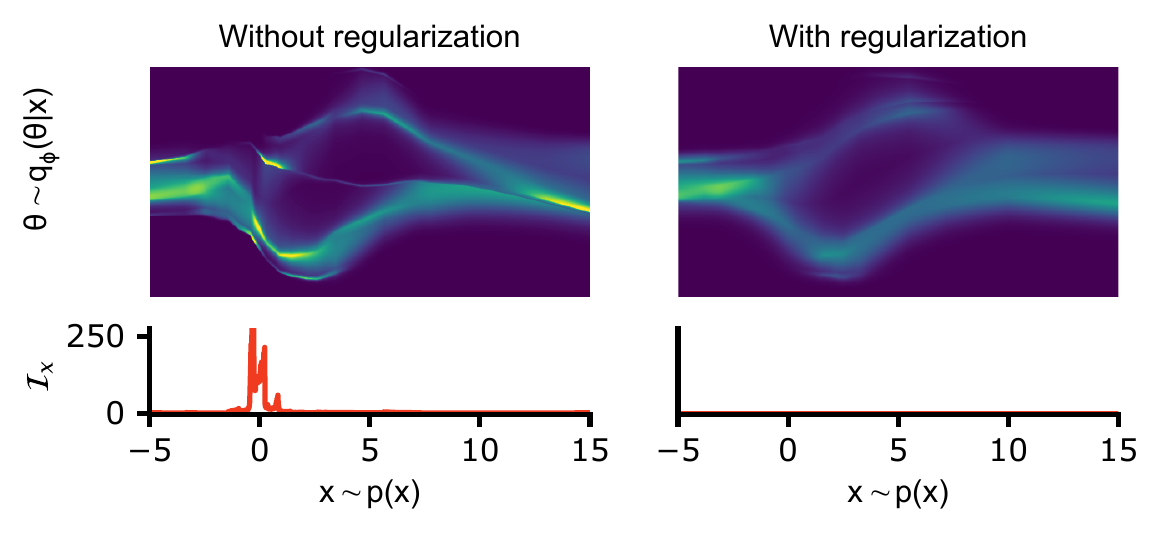}
    \caption{\textbf{Regularizing conditional density estimators by the Fisher information matrix (FIM)}. We trained a neural spline flow to estimate a conditional density with negative log-likelihood loss (left) and with our FIM regularizer (right). The Fisher information (bottom) is high in regions that are non-smooth along the conditioning variable $x$. The regularized loss leads to density estimates which are smoother while still being able to capture complex densities.}
    \label{fig:fisher_reg}
\end{figure}

We demonstrate the method on a simple one-dimensional conditional density estimation task using a neural spline flow \citep{dolatabadi2020invertible} (Fig.~\ref{fig:fisher_reg}). The Fisher information is large in $\vx$-regions where $q_\phi(\vtheta | \vx)$ changes quickly as a function of $\vx$. By regularizing with the trace of the FIM, the learned density is significantly smoother.

\begin{figure*}[t]
    \centering
    \includegraphics[width=\textwidth]{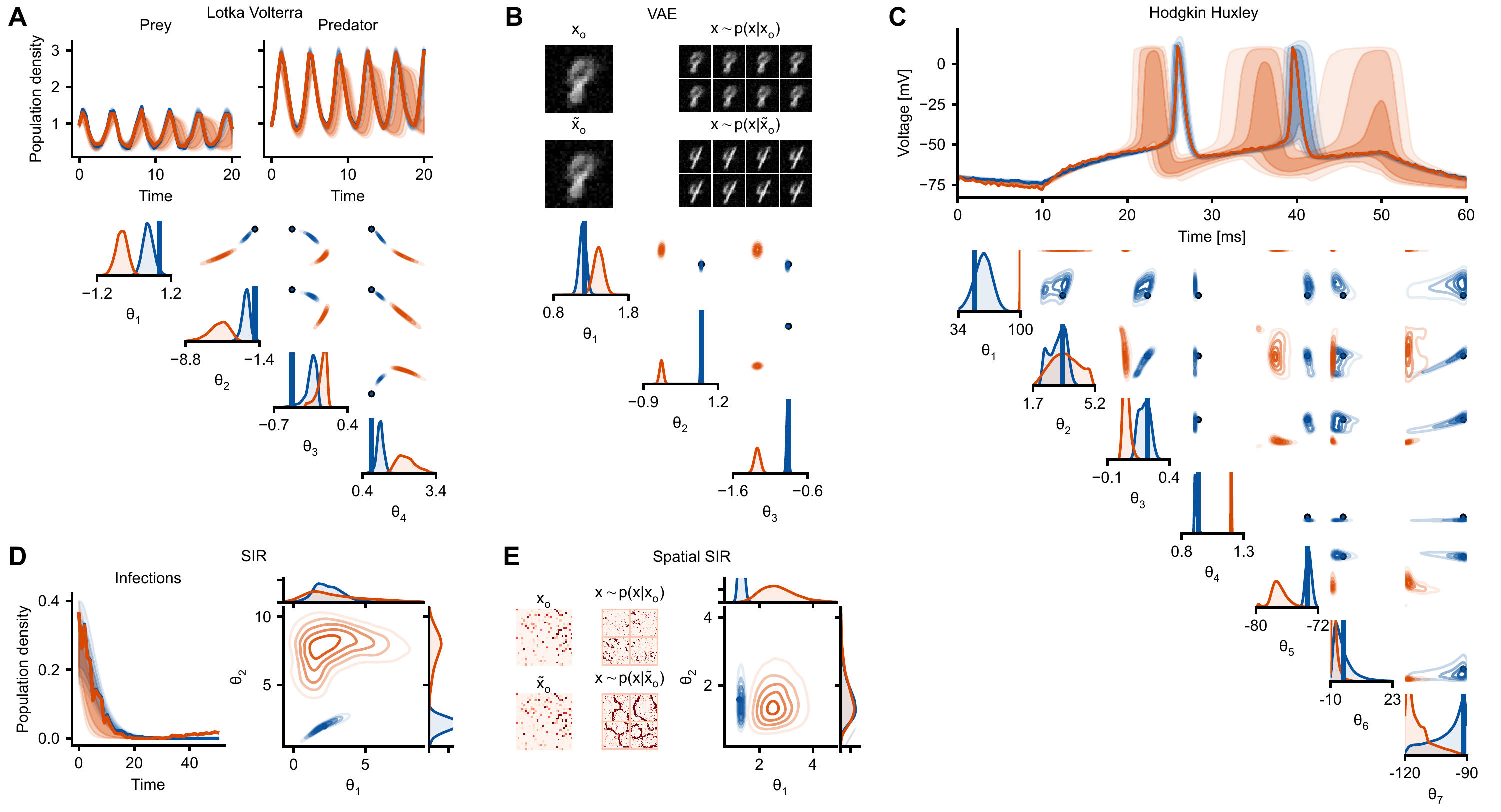}
    \caption{\textbf{Adversarial examples for each benchmark task}. Each panel shows i) the original observation (blue line) and corresponding posterior predictive samples (blue shaded), ii) the adversarial example (orange line) and posterior predictive samples based on the perturbed posterior estimate, and iii) posterior distribution plots with the posterior estimate for the original (blue) and perturbed (orange) data, and the ground-truth parameters (black dot).
    %posterior predictives and the posterior estimate. In the posterior predictive plots, the clean observation is shown as a blue line, its posterior predictive samples as blue shades, the adversarial example as an orange line, and the posterior predictive given the adversarial example as orange shades. In the posterior distribution plots, the posterior given clean data is in blue, the posterior given the adversarial example is in orange, and the black dot is the parameter set that generated the observation.
    }
    \label{fig:adversarial_examples}
\end{figure*}

%\subsection{Metrics for evaluating adversarial robustness}

%We investigated two metrics to study how strongly the adversarial attack influences the predicted posterior: First, the Kullback-Leibler divergence between the predicted posterior given clean and perturbed data $D_{KL}(q||\tilde{q})$ and, second, the expected coverage of the perturbed posterior estimate $\tilde{q}$.% allows us to evaluate not only the accuracy of the posterior approximation but also whether it is conservative or overconfident. 

\section{Experimental results}

%We examined the robustness of Neural Posterior Estimation (NPE), with and without defenses on benchmark tasks.
%as well as a challenging inference task on a pyloric network simulator \citep{prinz_similar_2004, deistler2022energy}.

\subsection{Benchmark tasks}

We first evaluated the robustness of Neural Posterior Estimation (NPE) and the effect of FIM-regularization on six benchmark tasks (details in %Appendix 
Sec.~\ref{sec:appendix_benchmark}). Rather than using established benchmark tasks \citep{lueckmann_benchmarking_2021}, we chose tasks with more high-dimensional data, which might offer more flexibility for adversarial attacks.

\paragraph{Visualizing adversarial attacks}

We first visualized the effect of several adversarial examples on inference models trained with standard (i.e., unregularized) NPE. We trained NPE with a Masked Autoregressive Flow (MAF, \citealt{papamakarios_masked_2017}) on $100k$ simulations and generated an adversarial attack for a held-out datapoint. Although the perturbations to the observations are hardly perceptible, the posterior estimates change drastically, and posterior predictive samples match neither the clean nor the perturbed observation (Fig.~\ref{fig:adversarial_examples}). This indicates that the attacked density estimator predicts a posterior distribution that does not match the true Bayesian posterior given the perturbed datapoint $p(\vtheta|\tilde{\vx})$, but rather it predicts an incorrect distribution.

How does the adversarial attack change the prediction of the neural density estimator so strongly? We investigated two possibilities for this: First, the adversarial attack could construct a datapoint $\tilde{\vx}$ which is misspecified. Previous work has reported that NPE can perform poorly in the presence of misspecification \citep{cannon_investigating_2022}. Indeed, on the SIR benchmark task (Fig.~\ref{fig:adversarial_examples}D), we find clues that are consistent with misspecification: At the end of the simulation ($t > 20$), the perturbed observation shows an increase in infections although they had already nearly reached zero. Such an increase cannot be modeled by the simulator and cannot be attributed to the noise model (since the noise is log-normal and, thus, small for low infection counts).

A second possibility for the adversarial attack to strongly change the posterior estimate would be to exploit the neural network itself and generate an attack for which the network produces poor predictions. We hypothesized that, on our benchmark tasks, this possibility would dominate. To investigate this, we performed adversarial attacks on different density estimators and evaluated how similar the adversarial attacks were to each other %(%Appendix 
(Fig.~\ref{fig:attack_thetas}). 
%If the constructed attacks were similar, this would suggest that the attack exploits properties in the simulator (e.g., by generating a misspecified example). 
We find that the attacks largely differ between different density estimators, suggesting that the attacks are indeed targeted to the specific neural network.
%Our results, presented in Figure \ref{fig:npe_eval}, not only affirm their findings but also reveal that adversarially chosen perturbations of the same magnitude can cause much larger errors. 
%Overall, these analyses show that adversarial attacks can leverage model misspecification and the vulnerability of neural networks.

\begin{figure*}[tp]
    \centering
    \includegraphics[width=\textwidth]{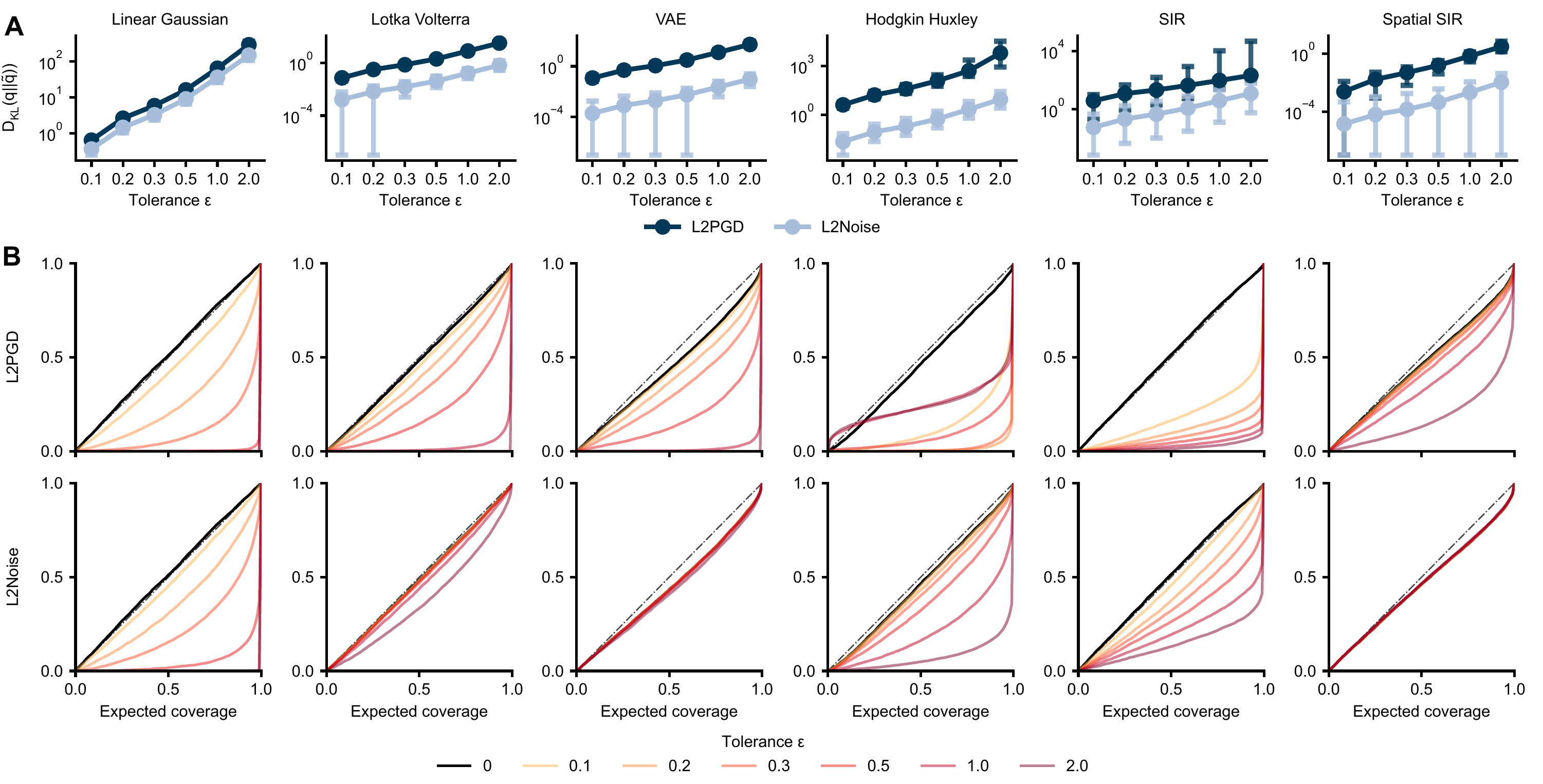}
    \caption{\textbf{Adversarial attacks on neural posterior estimation}.
    \textbf{(A)} KL-divergence between posterior estimates for original and perturbed data, $D_{KL}(q || \tilde{q})$ for targeted (L2PGD) and random (L2Noise) attacks on a linear Gaussian model and six benchmark tasks (details in %Appendix
    Sec.~\ref{sec:appendix_benchmark}), for several tolerance levels. Error bars show 15\% and 85\% quantiles. \textbf{(B)} Nominal coverage vs. empirical expected coverage for L2PGD (top) and L2Noise (bottom) attacks. The dotted line is identity.}
    \label{fig:attacks}
\end{figure*}

\paragraph{Quantifying the impact of adversarial attacks}

We quantified the effect of adversarial attacks on NPE without using an adversarial defense. 
After training NPE with 100k simulations, we constructed adversarial attacks for $10^4$ held-out datapoints (as described in Sec.~\ref{sec:adversarial_attacks}). As a baseline, we also added a random perturbation of the same magnitude on each datapoint.
We then computed the average $D_{KL}$ between the posterior estimates given clean and perturbed data (Fig.~\ref{fig:attacks}). For all tasks and tolerance levels (the scale of the perturbation), the adversarial attack increases the $D_{KL}$ more strongly than a random attack. In addition, for all tasks apart from the linear Gaussian task, the difference between the adversarial and the random attack is several orders of magnitude (Fig.~\ref{fig:attacks}A).

As a second evaluation-metric, we computed the expected coverage of the perturbed posterior estimates, which allows us to study whether posterior estimates are under-, or overconfident (Fig.~\ref{fig:attacks}B, details in Sec.~\ref{sec:appendix_metrics}) \citep{cannon_investigating_2022}. For stronger perturbations, the posterior estimates become overconfident around wrong parameter regions and show poor coverage. As expected, adversarial attacks impact the coverage substantially more strongly than random attacks.

Additional results for different density estimators, alternative attack definitions, and simulation budgets can be found in Sec. \ref{sec:attack_additional_results} (%Appendix 
Figs.~\ref{fig:fig3_appendix}, \ref{fig:coverages_per_density estimator}, \ref{fig:mmd_mle_attack}). The results are mostly consistent across different density estimators (with minor exceptions at low simulation budgets), indicating that more flexible estimators are not necessarily less robust.

%may not negatively impact robustness.% We note, however, that the $D_{KL}$ can be computed analytically some of our density estimators (whereas others rely on Monte Carlo techniques), which might make the attack more effective for simpler density estimators.

\begin{figure*}[t]
    \centering
    \includegraphics[width=\textwidth]{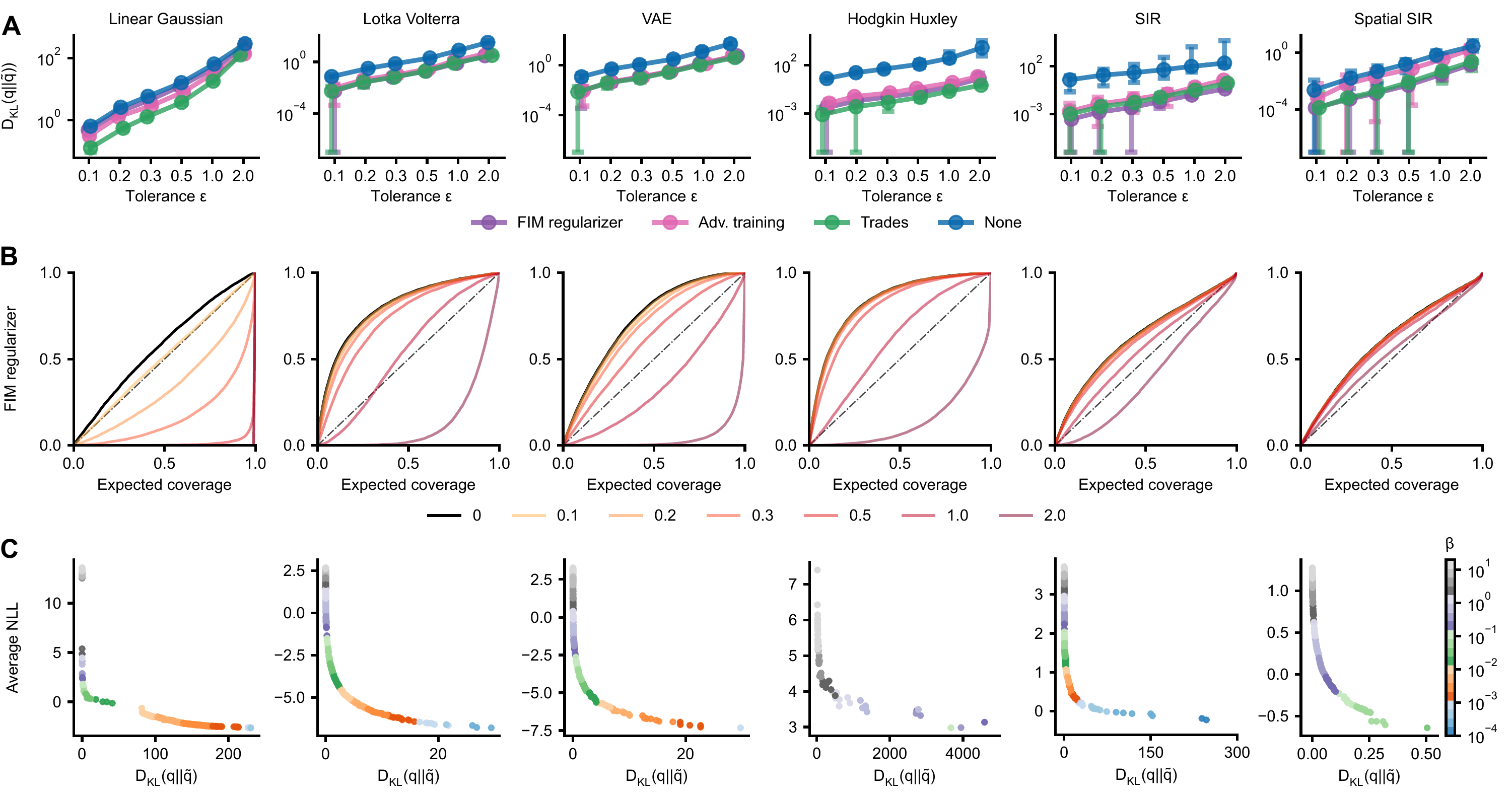}
    
    \caption{\textbf{Defenses against adversarial attacks}. \textbf{(A)} KL-divergences $D_{KL}(q || \tilde{q})$ for all three defenses (FIM regularisation, adversarial training and TRADES, see \ref{sec:appendix_defenses} for details) and without defense for six benchmark tasks. \textbf{(B)} Expected coverage for FIM regularization. \textbf{(C)} Trade-offs between accuracy (average of log-likelihood on unperturbed data) and robustness ($D_{KL}(q||\tilde{q})$) for $\epsilon=2.0$. For a range of regularisation strengths $\beta$, a large gain in robustness only leads to a small drop in accuracy.}
    \label{fig:defenses_attacked}
\end{figure*}

\paragraph{Adversarial defense of NPE}

Next, we evaluated the adversarial robustness when regularizing NPE with the moving average estimate of the trace of the Fisher Information Matrix (FIM) (Sec.~\ref{sec:adversarial_defense}). In addition, we evaluated two approaches adapted from defense methods for classification tasks-- however, both of these approaches rely on generating adversarial examples during training and are, thus, more computationally expensive (details in Sec.~\ref{appendix:defense_methods}, methods are labeled as `Adv.~training' and `TRADES'). 

All adversarial defense methods significantly reduce the ability of attacks to change the posterior estimate (Fig.~\ref{fig:defenses_attacked}A). In addition, the FIM regularizer performs similarly to other defense methods but is computationally much more efficient and scalable (\ref{sec:runtime}, Fig.~\ref{fig:runtime}, sweeps for $\beta$ in Fig.~\ref{fig:fim_hyperparameters}).

We evaluated the expected coverage when using FIM regularization (Fig.~\ref{fig:defenses_attacked}B, results for Adv.~Training and TRADES in Fig.~\ref{fig:appendix_defenses_adv_training_trades}). For all tasks, the coverage is shifted towards the upper left corner, indicating a more conservative posterior estimate (further analysis in Sec.~\ref{sec:appendix_tradeoff}). Even for medium to high tolerance levels (i.e., strong perturbations), the posterior estimate often remains underconfident and covers the true parameter set, a behavior which has been argued to be desirable in scientific applications \citep{hermans_trust_2022}. Other defense methods (that were not specifically developed as adversarial defenses), such as posterior ensembles or noise augmentation, barely increase the adversarial robustness of NPE (Fig.~\ref{fig:appendix_defenses_adv_training_trades}, Sec.~\ref{sec:appendix_defenses}). Further, we investigate this effect directly comparing against the true posterior (as estimated via MCMC for a subset of tasks) in Sec.~\ref{sec:appendix_true_post}, verifying that posterior approximation on adversarial perturbed data is poor but can be improved using FIM regularization.

Finally, we studied the trade-off between robustness to adversarial perturbations and accuracy of the posterior estimate on unperturbed data \citep{zhang_theoretically_2019, tsipras_robustness_2019}. We computed the accuracy on unperturbed data (evaluated as average log-likelihood) and the robustness to adversarial perturbations (measured as $D_{KL}$ between clean and perturbed posteriors) for a range of regularization strengths $\beta$ (Fig.~\ref{fig:defenses_attacked}C). For a set of intermediate values for $\beta$, it is possible to achieve a large gain in robustness while only weakly reducing accuracy (details in Sec.~\ref{sec:appendix_tradeoff}, results for other density estimators in Sec.~\ref{sec:appendix_defenses}, Figs.~\ref{fig:fim_hyperparameters} and \ref{fig:appendix_defenses_density_estimators}).

Overall, FIM regularization is a computationally efficient method to reduce the impact of adversarial examples on NPE. While it encourages underconfident posterior estimates, it allows for high robustness with a relatively modest reduction in accuracy.

\begin{figure*}[tp]
    \centering
    \includegraphics[width=\textwidth]{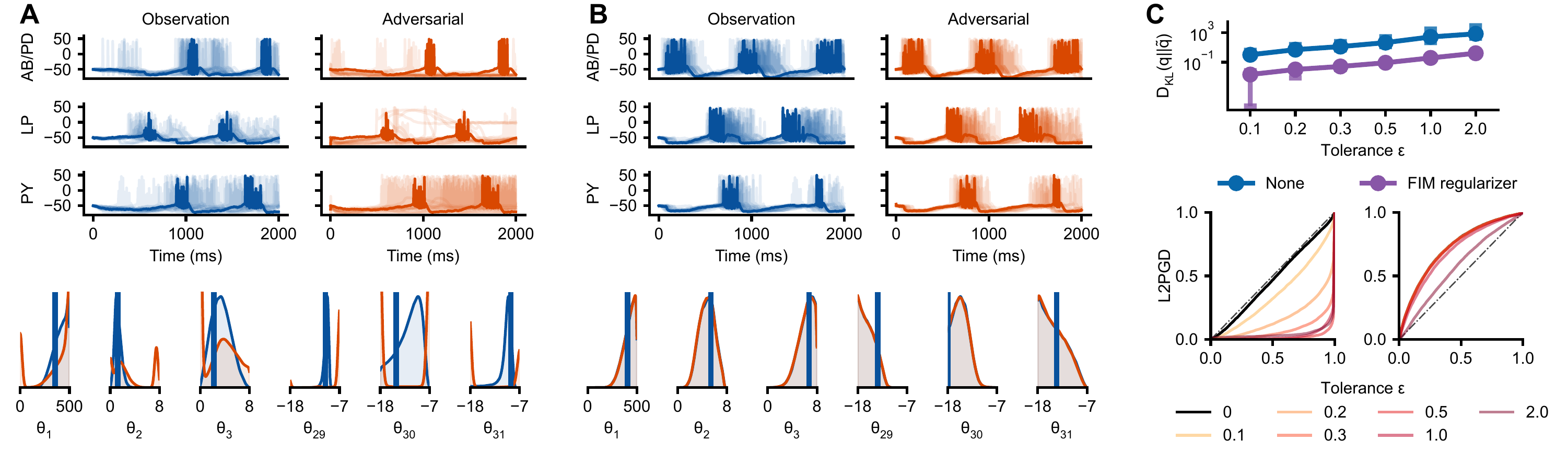}
    
    \caption{\textbf{Attack and defense on pyloric network simulator.} \textbf{(A)} Adversarial attack on NPE. Top: Observation (blue line), posterior predictives (blue shades), adversarially perturbed observation (orange line) and corresponding posterior predictives (orange shades). Bottom: Subset of marginals of posterior distribution given clean (blue) and perturbed (orange) data. Blue line is the true parameter set. \textbf{(B)} Same as A, but when employing FIM-regularization. \textbf{(C)} Top: $D_{KL}(q || \tilde{q})$ for NPE and FIM-regularized NPE. Bottom: Expected coverage for NPE (left) and FIM-regularized NPE (right).}
    \label{fig:pyloric}
\end{figure*}

\subsection{Neuroscience example: Pyloric network}
Finally, we performed adversarial attacks and defenses on a real-world simulator of the pyloric network in the stomatogastric ganglion (STG) of the crab \textit{Cancer Borealis}. The simulator includes three model neurons, each with eight membrane conductances and seven synapses (31 parameters in total) \citep{prinz2003alternative, prinz_similar_2004}. Prior studies have used extensive simulations from prior samples and performed amortized inference with NPE (18 million simulations in \citet{goncalves_training_2020}, 9 million in \citet{deistler2022energy}). Both of these studies used hand-crafted summary statistics. In contrast, we here performed inference on the raw traces (subsampled by a factor of 100 due to memory constraints).% by using a convolutional embedding network.

After subsampling, the data contains three voltage traces, each of length $800$. We ran 8M simulations from the prior and excluded any parameters which generated physiologically implausible data \citep{lueckmann_flexible_2017}, resulting in a dataset with $750k$ datapoints.
%This does not impact the accuracy of the posterior approximation as demonstrated in \cite{lueckmann_flexible_2017} for any $x$ that also satisfies the same criterion. We thus restrict also our evaluation to such $x$, which is known as calibration. 
We used a one-dimensional convolutional neural network for each of the three traces and passed the resulting embedding through a fully-connected neural network (Fig.~\ref{fig:pyloric}).

The neural density estimator, trained without regularization, is susceptible to adversarial attacks (Fig.~\ref{fig:pyloric}A). Given unperturbed data, the posterior predictive closely matches the data, whereas for the adversarially perturbed data the posterior predictive no longer matches the observations. In addition, the predicted posterior given clean data strongly differs from the predicted posterior given adversarially perturbed data.

In contrast, when regularizing with the FIM approach, the neural density estimator becomes significantly more robust to adversarial perturbations (Fig.~\ref{fig:pyloric}B). The posterior predictives now closely matched the data, both for clean as well as adversarially perturbed observations. In addition, the posterior estimates given clean and perturbed observations match closely.

We quantified these results by computing the $D_{KL}$ between clean and adversarially perturbed posterior estimates as well as the expected coverage (Fig.~\ref{fig:pyloric}C). NPE without regularization has a higher $D_{KL}$ and is overconfident. In contrast, the FIM-regularized posterior is underconfident, even for strong perturbations.
These results demonstrate that real-world simulators can strongly suffer from adversarial attacks. The results also show that our proposed FIM-regularizer scales to challenging and high-dimensional tasks.

\section{Discussion}

We showed that amortized Bayesian inference can be vulnerable to adversarial attacks. The posterior estimate can change strongly when slightly perturbing the observed data, leading to inaccurate inference results. This poses a difficult challenge for amortized Bayesian inference which would severely limit its utility for applications in which trustworthy posterior estimates are essential: If small changes in the input data can have a strong impact on inference results, using misspecified data, or simply data not encountered during training could also lead to severely wrong conclusions.

To address this issue, we propose a computationally efficient defense strategy that can be used to reduce the vulnerability of Neural Posterior Estimation to adversarial attacks. We demonstrate the effectiveness of this method and show that it can significantly improve the robustness and reliability of NPE in the presence of adversarial attacks. % which makes it a reliable method in real-world applications.% We presented a scalable defense method baesd on a moving average of the trace of the Fisher Information Matrix (FIM), and demonstrated its effectiveness on a challenging high-dimensional inference task.

\paragraph{Prior work on adversarial attacks and defenses} Adversarial attacks and defenses have been studied on variational autoencoders \citep{kuzina_alleviating_2022, husain_adversarial_2022, shu_amortized_2018, barrett_certifiably_2022, willetts_improving_2021, akrami_robust_2022}. Our work differs from these papers in that we focus on posterior distributions parameterized by expressive conditional density estimators such as normalizing flows. Note, crucially, that the attacks in this context are on the conditioning-variable, in contrast to previous work on flow-based models studying attacks on the \emph{output} of (unconditional) flow-based models \citep{pope_adversarial_2020}. To train our inference model, we use the negative log-likelihood as loss-function (as compared to the ELBO for variational autoencoders), which makes our approach applicable to non-differentiable and implicit models. %We focus on conditional flow-based models and adversarial perturbations of the conditioning variable, in contrast to previous work on flow-based models . 
%
%Prior work on model-misspecification for simulation-based inference has evaluated the sensitivity of NPE to untargeted perturbations\citep{cannon_investigating_2022}.  %i.e., attacks that do not exploit the neural network parameters of NPE . 
%In contrast, we evaluate adversarial robustness on a wide range of models and tasks.
%
%\paragraph{Adversarial defense:} 
%Methods for certifying robustness of generative models have been proposed \citep{barrett_certifiably_2022, willetts_improving_2021, akrami_robust_2022}, but they are often limited to Gaussian inference models. \citet{shu_amortized_2018} suggest adding noise to improve generalization in VAEs, but do not consider adversarial attacks. Notably, our defense is related to input gradient regularization schemes that showed scalable success within the classification framework \cite{finlay_scaleable_2019}.
%Detection of model misspecification is proposed in \citet{schmitt_detecting_2022}, but detection methods in adversarial setting are often unreliable \cite{carlini_adversarial_2017}. 
Several recent studies have proposed improvements to the robustness of NPE \citep{dellaporta_robust_2022,lemos_robust_2022,ward_robust_2022, matsubara_robust_2022, finlay_scaleable_2019}, but none of them have considered defenses against adversarial attacks. 
\citet{dax_neural_2022} proposed the use of (likelihood-based) importance sampling to identify and correct poor approximations in an application from astrophysics, and in this context also evaluated the adversarial robustness of a neural posterior estimator. Concurrent theoretical work of \citet{altekruger2023conditional} established basic conditions under which conditional density estimators on convergence are provably robust in particular depending on the Lipschitz constant of the inference network with respect to the observation.

\paragraph{Amortized Bayesian inference}
We studied adversarial robustness of a particular amortized Bayesian inference algorithm, Neural Posterior Estimation (NPE). Other simulation-based inference methods can be categorized as amortized as well, e.g., Neural Ratio Estimation (NRE) or Neural Likelihood Estimation (NLE) \citep{cranmer2020frontier,hermans_likelihood-free_2020, papamakarios_sequential_2019}. These methods do not require new simulations or network training for new observations, but they require a (potentially expensive) inference phase to obtain the posterior. In addition, another approach to amortized Bayesian inference would be to perform amortized variational inference, which requires a differentiable model and likelihood-evaluations. We leave the study of adversarial attacks in these methods to future work.

\paragraph{Model misspecification and adversarial robustness}
Previous work \citep{cannon_investigating_2022} raised concerns about the reliability of NPE on misspecified simulators. %They primarily attribute the issues of NPE in misspecified models to the insufficient out-of-distribution generalizability of the networks. 
Adversarial examples exploit the brittleness of neural networks to construct examples on which NPE performs particularly poorly. As such, our study can be considered as a worst-case scenario of how minor deviations in the observed data can impact its reliability. We find that adversarial examples depend strongly on the network (for the same simulator), indicating the crucial role of the inference network.
%are as minor deviations from well-specified data, yet they can still have a significant impact on the performance of neural networks.
%This highlights a fundamental issue with the brittleness of these networks, which is not solely caused by misspecification, but also by the limitations of current training algorithms and neural networks. 

\paragraph{Limitations}
Using a defense against adversarial attacks comes at two costs: Increased computational cost and, potentially, broader posteriors on clean data. Our proposed regularization scheme largely reduces computational cost (by up to an order of magnitude compared to other defense methods such as TRADES), but it, nonetheless, requires drawing Monte Carlo samples from the posterior estimate and evaluating its gradient w.r.t.~every datapoint at every epoch. Across six benchmark tasks, our regularizer increased training time by a factor of four (compared to standard NPE, Fig.~\ref{fig:runtime}).
 Our analysis places emphasis on inference networks that have the capability to learn summary statistics of complex data, if necessary, in an end-to-end manner using neural networks \citep{Chan2018-cw, radev_bayesflow_2020}. However, in various applications, expert-crafted summary statistics are commonly employed which can be explicitly designed to be robust against certain perturbations.

It has been argued that, in many applications (and in particular in the natural sciences), it is desirable to have underconfident posteriors \cite{hermans_trust_2022} -- however, posterior estimates that are systematically too broad lead to a lower rate of learning from data and, thus, slower information acquisition. While we demonstrated that our method is comparable to NPE in terms of negative log-likelihood, users might need to evaluate the trade-off between robustness and information acquisition for their applications.

% metrics - currently only uses the distance to true posterior

%\paragraph{Conclusion}
%We studied the robustness of amortized Bayesian inference to adversarial attacks. We demonstrated that small, almost imperceptible changes to the data can lead to dramatically different posterior approximations. We then suggested a computationally cheap method to improve the robustness of NPE and demonstrated its efficacy on six benchmark tasks and a real-world example from neuroscience. 

% \textbf{Do not} include acknowledgements in the initial version of
% the paper submitted for blind review.
%acknowledge SImaelsam !

\section*{Acknowledgements}

We thank Pedro Gon{c}alves, Jaivardhan Kapoor, and Maximilian Dax for discussions and comments on the manuscript.
MG and MD are supported by the International Max Planck Research School for Intelligent Systems (IMPRS-IS).
This work was supported by the German Research Foundation (DFG) through Germany’s Excellence Strategy – EXC-Number 2064/1 – Project number 390727645, the German Federal Ministry of Education and Research (Tübingen AI Center, FKZ: 01IS18039A), and the `Certification and Foundations of Safe Machine Learning Systems in Healthcare' project funded by the Carl Zeiss Foundation.

\section*{Software and Data}
We used PyTorch for all neural networks \cite{paszke_pytorch_2019} and hydra to track all configurations \citep{Yadan2019Hydra}. Code to reproduce results is available at \url{https://github.com/mackelab/RABI}.

% Acknowledgements should only appear in the accepted version.
%\section*{Acknowledgements}

%\textbf{Do not} include acknowledgements in the initial version of
%the paper submitted for blind review.

\bibliography{main}

\begin{thebibliography}{77}
\providecommand{\natexlab}[1]{#1}
\providecommand{\url}[1]{\texttt{#1}}
\expandafter\ifx\csname urlstyle\endcsname\relax
  \providecommand{\doi}[1]{doi: #1}\else
  \providecommand{\doi}{doi: \begingroup \urlstyle{rm}\Url}\fi

\bibitem[Akrami et~al.(2022)Akrami, Joshi, Li, Aydöre, and
  Leahy]{akrami_robust_2022}
Akrami, H., Joshi, A.~A., Li, J., Aydöre, S., and Leahy, R.~M.
\newblock A robust variational autoencoder using beta divergence.
\newblock \emph{Knowledge-Based Systems}, 238:\penalty0 107886, 2022.
\newblock Publisher: Elsevier.

\bibitem[Altekr{\"u}ger et~al.(2023)Altekr{\"u}ger, Hagemann, and
  Steidl]{altekruger2023conditional}
Altekr{\"u}ger, F., Hagemann, P., and Steidl, G.
\newblock Conditional generative models are provably robust: Pointwise
  guarantees for bayesian inverse problems.
\newblock \emph{arXiv preprint arXiv:2303.15845}, 2023.

\bibitem[Andrieu \& Thoms(2008)Andrieu and Thoms]{andrieu2008tutorial}
Andrieu, C. and Thoms, J.
\newblock A tutorial on adaptive mcmc.
\newblock \emph{Statistics and computing}, 18:\penalty0 343--373, 2008.

\bibitem[Arnst et~al.(2022)Arnst, Louppe, Van~Hulle, Gillet, Bureau, and
  Denoël]{arnst_hybrid_2022}
Arnst, M., Louppe, G., Van~Hulle, R., Gillet, L., Bureau, F., and Denoël, V.
\newblock A hybrid stochastic model and its {Bayesian} identification for
  infectious disease screening in a university campus with application to
  massive {COVID}-19 screening at the {University} of {Liège}.
\newblock \emph{Mathematical Biosciences}, 347:\penalty0 108805, May 2022.

\bibitem[Barrett et~al.(2022)Barrett, Camuto, Willetts, and
  Rainforth]{barrett_certifiably_2022}
Barrett, B., Camuto, A., Willetts, M., and Rainforth, T.
\newblock Certifiably robust variational autoencoders.
\newblock In \emph{International {Conference} on {Artificial} {Intelligence}
  and {Statistics}}, pp.\  3663--3683. PMLR, 2022.

\bibitem[Berger et~al.(1994)Berger, Moreno, Pericchi, Bayarri, Bernardo, Cano,
  De~la Horra, Martín, Ríos-Insúa, Betrò, Dasgupta, Gustafson, Wasserman,
  Kadane, Srinivasan, Lavine, O’Hagan, Polasek, Robert, Goutis, Ruggeri,
  Salinetti, and Sivaganesan]{berger_overview_1994}
Berger, J.~O., Moreno, E., Pericchi, L.~R., Bayarri, M.~J., Bernardo, J.~M.,
  Cano, J.~A., De~la Horra, J., Martín, J., Ríos-Insúa, D., Betrò, B.,
  Dasgupta, A., Gustafson, P., Wasserman, L., Kadane, J.~B., Srinivasan, C.,
  Lavine, M., O’Hagan, A., Polasek, W., Robert, C.~P., Goutis, C., Ruggeri,
  F., Salinetti, G., and Sivaganesan, S.
\newblock An overview of robust {Bayesian} analysis.
\newblock \emph{Test}, 3\penalty0 (1):\penalty0 5--124, June 1994.

\bibitem[Bingham et~al.(2019)Bingham, Chen, Jankowiak, Obermeyer, Pradhan,
  Karaletsos, Singh, Szerlip, Horsfall, and Goodman]{bingham_pyro_2019}
Bingham, E., Chen, J.~P., Jankowiak, M., Obermeyer, F., Pradhan, N.,
  Karaletsos, T., Singh, R., Szerlip, P., Horsfall, P., and Goodman, N.~D.
\newblock Pyro: {Deep} {Universal} {Probabilistic} {Programming}.
\newblock \emph{Journal of Machine Learning Research}, 20\penalty0
  (28):\penalty0 1--6, 2019.
\newblock ISSN 1533-7928.

\bibitem[Blyth(1994)]{blyth_local_1994}
Blyth, S.
\newblock Local {Divergence} and {Association}.
\newblock \emph{Biometrika}, 81\penalty0 (3):\penalty0 579--584, 1994.
\newblock Publisher: [Oxford University Press, Biometrika Trust].

\bibitem[Boelts et~al.(2022)Boelts, Lueckmann, Gao, and
  Macke]{boelts_flexible_2022}
Boelts, J., Lueckmann, J.-M., Gao, R., and Macke, J.~H.
\newblock Flexible and efficient simulation-based inference for models of
  decision-making.
\newblock \emph{eLife}, 11:\penalty0 e77220, July 2022.

\bibitem[Cannon et~al.(2022)Cannon, Ward, and
  Schmon]{cannon_investigating_2022}
Cannon, P., Ward, D., and Schmon, S.~M.
\newblock Investigating the {Impact} of {Model} {Misspecification} in {Neural}
  {Simulation}-based {Inference}, September 2022.
\newblock arXiv:2209.01845 [cs, stat].

\bibitem[Chan et~al.(2018)Chan, Perrone, Spence, Jenkins, Mathieson, and
  Song]{Chan2018-cw}
Chan, J., Perrone, V., Spence, J.~P., Jenkins, P.~A., Mathieson, S., and Song,
  Y.~S.
\newblock A {Likelihood-Free} inference framework for population genetic data
  using exchangeable neural networks.
\newblock \emph{Adv Neural Inf Process Syst}, 31:\penalty0 8594--8605, December
  2018.

\bibitem[Cranmer et~al.(2020)Cranmer, Brehmer, and Louppe]{cranmer2020frontier}
Cranmer, K., Brehmer, J., and Louppe, G.
\newblock The frontier of simulation-based inference.
\newblock \emph{Proceedings of the National Academy of Sciences}, 117\penalty0
  (48):\penalty0 30055--30062, 2020.

\bibitem[Croce et~al.(2021)Croce, Andriushchenko, Sehwag, Debenedetti,
  Flammarion, Chiang, Mittal, and Hein]{croce_robustbench_2021}
Croce, F., Andriushchenko, M., Sehwag, V., Debenedetti, E., Flammarion, N.,
  Chiang, M., Mittal, P., and Hein, M.
\newblock Robustbench: a standardized adversarial robustness benchmark.
\newblock In \emph{Thirty-fifth Conference on Neural Information Processing
  Systems Datasets and Benchmarks Track (Round 2)}, 2021.

\bibitem[Dang-Nhu et~al.(2020)Dang-Nhu, Singh, Bielik, and
  Vechev]{dang-nhu_adversarial_2020}
Dang-Nhu, R., Singh, G., Bielik, P., and Vechev, M.
\newblock Adversarial attacks on probabilistic autoregressive forecasting
  models.
\newblock In III, H.~D. and Singh, A. (eds.), \emph{Proceedings of the 37th
  International Conference on Machine Learning}, volume 119 of
  \emph{Proceedings of Machine Learning Research}, pp.\  2356--2365. PMLR,
  13--18 Jul 2020.

\bibitem[Dax et~al.(2021)Dax, Green, Gair, Macke, Buonanno, and
  Sch{\"o}lkopf]{dax2021real}
Dax, M., Green, S.~R., Gair, J., Macke, J.~H., Buonanno, A., and Sch{\"o}lkopf,
  B.
\newblock Real-time gravitational wave science with neural posterior
  estimation.
\newblock \emph{Physical review letters}, 127\penalty0 (24):\penalty0 241103,
  2021.

\bibitem[Dax et~al.(2022)Dax, Green, Gair, Pürrer, Wildberger, Macke,
  Buonanno, and Schölkopf]{dax_neural_2022}
Dax, M., Green, S.~R., Gair, J., Pürrer, M., Wildberger, J., Macke, J.~H.,
  Buonanno, A., and Schölkopf, B.
\newblock Neural {Importance} {Sampling} for {Rapid} and {Reliable}
  {Gravitational}-{Wave} {Inference}, October 2022.
\newblock arXiv:2210.05686 [astro-ph, physics:gr-qc].

\bibitem[Deistler et~al.(2022{\natexlab{a}})Deistler, Goncalves, and
  Macke]{deistler2022truncated}
Deistler, M., Goncalves, P.~J., and Macke, J.~H.
\newblock Truncated proposals for scalable and hassle-free simulation-based
  inference.
\newblock In Oh, A.~H., Agarwal, A., Belgrave, D., and Cho, K. (eds.),
  \emph{Advances in Neural Information Processing Systems}, 2022{\natexlab{a}}.

\bibitem[Deistler et~al.(2022{\natexlab{b}})Deistler, Macke, and
  Gon{\c{c}}alves]{deistler2022energy}
Deistler, M., Macke, J.~H., and Gon{\c{c}}alves, P.~J.
\newblock Energy-efficient network activity from disparate circuit parameters.
\newblock \emph{Proceedings of the National Academy of Sciences}, 119\penalty0
  (44):\penalty0 e2207632119, 2022{\natexlab{b}}.

\bibitem[Dellaporta et~al.(2022)Dellaporta, Knoblauch, Damoulas, and
  Briol]{dellaporta_robust_2022}
Dellaporta, C., Knoblauch, J., Damoulas, T., and Briol, F.-X.
\newblock Robust {Bayesian} {Inference} for {Simulator}-based {Models} via the
  {MMD} {Posterior} {Bootstrap}.
\newblock In \emph{Proceedings of {The} 25th {International} {Conference} on
  {Artificial} {Intelligence} and {Statistics}}, pp.\  943--970. PMLR, May
  2022.
\newblock ISSN: 2640-3498.

\bibitem[Dolatabadi et~al.(2020)Dolatabadi, Erfani, and
  Leckie]{dolatabadi2020invertible}
Dolatabadi, H.~M., Erfani, S., and Leckie, C.
\newblock Invertible generative modeling using linear rational splines.
\newblock In \emph{International Conference on Artificial Intelligence and
  Statistics}, pp.\  4236--4246. PMLR, 2020.

\bibitem[Finlay \& Oberman(2019)Finlay and Oberman]{finlay_scaleable_2019}
Finlay, C. and Oberman, A.~M.
\newblock Scaleable input gradient regularization for adversarial robustness,
  October 2019.
\newblock arXiv:1905.11468 [cs, stat].

\bibitem[Golub et~al.(1999)Golub, Hansen, and O'Leary]{golub1999tikhonov}
Golub, G.~H., Hansen, P.~C., and O'Leary, D.~P.
\newblock Tikhonov regularization and total least squares.
\newblock \emph{SIAM journal on matrix analysis and applications}, 21\penalty0
  (1):\penalty0 185--194, 1999.

\bibitem[Gondim-Ribeiro et~al.(2018)Gondim-Ribeiro, Tabacof, and
  Valle]{gondim-ribeiro_adversarial_2018}
Gondim-Ribeiro, G., Tabacof, P., and Valle, E.
\newblock Adversarial attacks on variational autoencoders.
\newblock \emph{arXiv preprint arXiv:1806.04646}, 2018.

\bibitem[Gonçalves et~al.(2020)Gonçalves, Lueckmann, Deistler, Nonnenmacher,
  Öcal, Bassetto, Chintaluri, Podlaski, Haddad, and
  Vogels]{goncalves_training_2020}
Gonçalves, P.~J., Lueckmann, J.-M., Deistler, M., Nonnenmacher, M., Öcal, K.,
  Bassetto, G., Chintaluri, C., Podlaski, W.~F., Haddad, S.~A., and Vogels,
  T.~P.
\newblock Training deep neural density estimators to identify mechanistic
  models of neural dynamics.
\newblock \emph{Elife}, 9:\penalty0 e56261, 2020.

\bibitem[Goodfellow et~al.(2015)Goodfellow, Shlens, and
  Szegedy]{goodfellow_explaining_2015}
Goodfellow, I.~J., Shlens, J., and Szegedy, C.
\newblock Explaining and {Harnessing} {Adversarial} {Examples}.
\newblock In \emph{International Conference on Learning Representations}, 2015.

\bibitem[Greenberg et~al.(2019)Greenberg, Nonnenmacher, and
  Macke]{greenberg_automatic_2019}
Greenberg, D., Nonnenmacher, M., and Macke, J.
\newblock Automatic posterior transformation for likelihood-free inference.
\newblock In \emph{International {Conference} on {Machine} {Learning}}, pp.\
  2404--2414. PMLR, 2019.

\bibitem[Gretton et~al.(2012)Gretton, Sejdinovic, Strathmann, Balakrishnan,
  Pontil, Fukumizu, and Sriperumbudur]{gretton_optimal_2012}
Gretton, A., Sejdinovic, D., Strathmann, H., Balakrishnan, S., Pontil, M.,
  Fukumizu, K., and Sriperumbudur, B.~K.
\newblock Optimal kernel choice for large-scale two-sample tests.
\newblock In \emph{Advances in {Neural} {Information} {Processing} {Systems}},
  volume~25. Curran Associates, Inc., 2012.

\bibitem[Grünwald \& Ommen(2017)Grünwald and
  Ommen]{grunwald_inconsistency_2017}
Grünwald, P. and Ommen, T.~v.
\newblock Inconsistency of {Bayesian} {Inference} for {Misspecified} {Linear}
  {Models}, and a {Proposal} for {Repairing} {It}.
\newblock \emph{Bayesian Analysis}, 12\penalty0 (4):\penalty0 1069--1103,
  December 2017.
\newblock Publisher: International Society for Bayesian Analysis.

\bibitem[Hayashi et~al.(2018)Hayashi, Yuan, and Liang]{hayashi_bias_2018}
Hayashi, K., Yuan, K.-H., and Liang, L.
\newblock On the {Bias} in {Eigenvalues} of {Sample} {Covariance} {Matrix}.
\newblock In Wiberg, M., Culpepper, S., Janssen, R., González, J., and
  Molenaar, D. (eds.), \emph{Quantitative {Psychology}}, Springer {Proceedings}
  in {Mathematics} \& {Statistics}, pp.\  221--233, Cham, 2018. Springer
  International Publishing.

\bibitem[Hayou(2017)]{hayou_overestimation_2017}
Hayou, S.
\newblock On the overestimation of the largest eigenvalue of a covariance
  matrix, August 2017.
\newblock arXiv:1708.03551 [math, q-fin, stat].

\bibitem[Hermans et~al.(2020)Hermans, Begy, and
  Louppe]{hermans_likelihood-free_2020}
Hermans, J., Begy, V., and Louppe, G.
\newblock Likelihood-free mcmc with amortized approximate ratio estimators.
\newblock In \emph{International {Conference} on {Machine} {Learning}}, pp.\
  4239--4248. PMLR, 2020.

\bibitem[Hermans et~al.(2022)Hermans, Delaunoy, Rozet, Wehenkel, Begy, and
  Louppe]{hermans_trust_2022}
Hermans, J., Delaunoy, A., Rozet, F., Wehenkel, A., Begy, V., and Louppe, G.
\newblock A {Trust} {Crisis} {In} {Simulation}-{Based} {Inference}? {Your}
  {Posterior} {Approximations} {Can} {Be} {Unfaithful}, December 2022.
\newblock arXiv:2110.06581 [cs, stat].

\bibitem[Holden et~al.(2009)Holden, Hauge, and Holden]{holden2009adaptive}
Holden, L., Hauge, R., and Holden, M.
\newblock {Adaptive independent Metropolis–Hastings}.
\newblock \emph{The Annals of Applied Probability}, 19\penalty0 (1):\penalty0
  395 -- 413, 2009.
\newblock \doi{10.1214/08-AAP545}.

\bibitem[Husain \& Knoblauch(2022)Husain and
  Knoblauch]{husain_adversarial_2022}
Husain, H. and Knoblauch, J.
\newblock Adversarial {Interpretation} of {Bayesian} {Inference}.
\newblock In \emph{Proceedings of {The} 33rd {International} {Conference} on
  {Algorithmic} {Learning} {Theory}}, pp.\  553--572. PMLR, March 2022.
\newblock ISSN: 2640-3498.

\bibitem[Kingma \& Welling(2014)Kingma and Welling]{kingma_auto-encoding_2022}
Kingma, D.~P. and Welling, M.
\newblock Auto-{Encoding} {Variational} {Bayes}.
\newblock In \emph{International Conference on Learning Representations}, 2014.

\bibitem[Kuzina et~al.(2022)Kuzina, Welling, and
  Tomczak]{kuzina_alleviating_2022}
Kuzina, A., Welling, M., and Tomczak, J.~M.
\newblock Alleviating adversarial attacks on variational autoencoders with
  {MCMC}.
\newblock In Oh, A.~H., Agarwal, A., Belgrave, D., and Cho, K. (eds.),
  \emph{Advances in Neural Information Processing Systems}, 2022.

\bibitem[Latz(2020)]{latz_well-posedness_2020}
Latz, J.
\newblock On the well-posedness of {Bayesian} inverse problems.
\newblock \emph{SIAM/ASA Journal on Uncertainty Quantification}, 8\penalty0
  (1):\penalty0 451--482, January 2020.
\newblock arXiv:1902.10257 [cs, math, stat].

\bibitem[Lemos et~al.(2022)Lemos, Cranmer, Abidi, Hahn, Eickenberg, Massara,
  Yallup, and Ho]{lemos_robust_2022}
Lemos, P., Cranmer, M., Abidi, M., Hahn, C., Eickenberg, M., Massara, E.,
  Yallup, D., and Ho, S.
\newblock Robust {Simulation}-{Based} {Inference} in {Cosmology} with
  {Bayesian} {Neural} {Networks}.
\newblock In \emph{ICML 2022 Workshop on Machine Learning for Astrophysics},
  2022.
\newblock arXiv:2207.08435 [astro-ph].

\bibitem[Li et~al.(2022)Li, Cheng, Hsieh, and Lee]{li_review_2022}
Li, Y., Cheng, M., Hsieh, C.-J., and Lee, T. C.~M.
\newblock A {Review} of {Adversarial} {Attack} and {Defense} for
  {Classification} {Methods}.
\newblock \emph{The American Statistician}, 76\penalty0 (4):\penalty0 329--345,
  October 2022.
\newblock arXiv:2111.09961 [cs].

\bibitem[Lueckmann et~al.(2017)Lueckmann, Goncalves, Bassetto, Öcal,
  Nonnenmacher, and Macke]{lueckmann_flexible_2017}
Lueckmann, J.-M., Goncalves, P.~J., Bassetto, G., Öcal, K., Nonnenmacher, M.,
  and Macke, J.~H.
\newblock Flexible statistical inference for mechanistic models of neural
  dynamics.
\newblock \emph{Advances in neural information processing systems}, 30, 2017.

\bibitem[Lueckmann et~al.(2021)Lueckmann, Boelts, Greenberg, Goncalves, and
  Macke]{lueckmann_benchmarking_2021}
Lueckmann, J.-M., Boelts, J., Greenberg, D., Goncalves, P., and Macke, J.
\newblock Benchmarking simulation-based inference.
\newblock In \emph{International {Conference} on {Artificial} {Intelligence}
  and {Statistics}}, pp.\  343--351. PMLR, 2021.

\bibitem[Madry et~al.(2018)Madry, Makelov, Schmidt, Tsipras, and
  Vladu]{madry_towards_2019}
Madry, A., Makelov, A., Schmidt, L., Tsipras, D., and Vladu, A.
\newblock Towards deep learning models resistant to adversarial attacks.
\newblock In \emph{International Conference on Learning Representations}, 2018.

\bibitem[Matsubara et~al.(2022)Matsubara, Knoblauch, Briol, and
  Oates]{matsubara_robust_2022}
Matsubara, T., Knoblauch, J., Briol, F.-X., and Oates, C.~J.
\newblock Robust {Generalised} {Bayesian} {Inference} for {Intractable}
  {Likelihoods}, January 2022.
\newblock arXiv:2104.07359 [math, stat].

\bibitem[Medina et~al.(2022)Medina, Olea, Rush, and
  Velez]{medina_robustness_2021}
Medina, M.~A., Olea, J. L.~M., Rush, C., and Velez, A.
\newblock On the robustness to misspecification of Î±-posteriors and their
  variational approximations.
\newblock \emph{Journal of Machine Learning Research}, 23\penalty0
  (147):\penalty0 1--51, 2022.

\bibitem[Min et~al.(2021)Min, Chen, and Karbasi]{min_curious_2020}
Min, Y., Chen, L., and Karbasi, A.
\newblock The curious case of adversarially robust models: More data can help,
  double descend, or hurt generalization.
\newblock In de~Campos, C. and Maathuis, M.~H. (eds.), \emph{Proceedings of the
  Thirty-Seventh Conference on Uncertainty in Artificial Intelligence}, volume
  161 of \emph{Proceedings of Machine Learning Research}, pp.\  129--139. PMLR,
  27--30 Jul 2021.

\bibitem[Miyato et~al.(2016)Miyato, Maeda, Koyama, Nakae, and
  Ishii]{miyato_distributional_2016}
Miyato, T., Maeda, S.-i., Koyama, M., Nakae, K., and Ishii, S.
\newblock Distributional {Smoothing} with {Virtual} {Adversarial} {Training}.
\newblock In \emph{International Conference on Learning Representations}, 2016.

\bibitem[Moon et~al.(2023)Moon, Oulasvirta, and Lee]{moon2023amortized}
Moon, H.-S., Oulasvirta, A., and Lee, B.
\newblock Amortized inference with user simulations.
\newblock In \emph{Proceedings of the 2023 CHI Conference on Human Factors in
  Computing Systems}, pp.\  1--20, 2023.

\bibitem[Neal(2003)]{neal2003slice}
Neal, R.~M.
\newblock Slice sampling.
\newblock \emph{The annals of statistics}, 31\penalty0 (3):\penalty0 705--767,
  2003.

\bibitem[Papamakarios \& Murray(2016)Papamakarios and
  Murray]{papamakarios_fast_2016}
Papamakarios, G. and Murray, I.
\newblock Fast $\epsilon$-free inference of simulation models with bayesian
  conditional density estimation.
\newblock \emph{Advances in neural information processing systems}, 29, 2016.

\bibitem[Papamakarios et~al.(2017)Papamakarios, Pavlakou, and
  Murray]{papamakarios_masked_2017}
Papamakarios, G., Pavlakou, T., and Murray, I.
\newblock Masked autoregressive flow for density estimation.
\newblock \emph{Advances in neural information processing systems}, 30, 2017.

\bibitem[Papamakarios et~al.(2019)Papamakarios, Sterratt, and
  Murray]{papamakarios_sequential_2019}
Papamakarios, G., Sterratt, D., and Murray, I.
\newblock Sequential neural likelihood: {Fast} likelihood-free inference with
  autoregressive flows.
\newblock In \emph{The 22nd {International} {Conference} on {Artificial}
  {Intelligence} and {Statistics}}, pp.\  837--848. PMLR, 2019.

\bibitem[Paszke et~al.(2019)Paszke, Gross, Massa, Lerer, Bradbury, Chanan,
  Killeen, Lin, Gimelshein, Antiga, Desmaison, Köpf, Yang, DeVito, Raison,
  Tejani, Chilamkurthy, Steiner, Fang, Bai, and Chintala]{paszke_pytorch_2019}
Paszke, A., Gross, S., Massa, F., Lerer, A., Bradbury, J., Chanan, G., Killeen,
  T., Lin, Z., Gimelshein, N., Antiga, L., Desmaison, A., Köpf, A., Yang, E.,
  DeVito, Z., Raison, M., Tejani, A., Chilamkurthy, S., Steiner, B., Fang, L.,
  Bai, J., and Chintala, S.
\newblock {PyTorch}: {An} {Imperative} {Style}, {High}-{Performance} {Deep}
  {Learning} {Library}, December 2019.
\newblock arXiv:1912.01703 [cs, stat].

\bibitem[Pope et~al.(2020)Pope, Balaji, and Feizi]{pope_adversarial_2020}
Pope, P., Balaji, Y., and Feizi, S.
\newblock Adversarial robustness of flow-based generative models.
\newblock In \emph{International {Conference} on {Artificial} {Intelligence}
  and {Statistics}}, pp.\  3795--3805. PMLR, 2020.

\bibitem[Pospischil et~al.(2008)Pospischil, Toledo-Rodriguez, Monier,
  Piwkowska, Bal, Frégnac, Markram, and Destexhe]{pospischil_minimal_2008}
Pospischil, M., Toledo-Rodriguez, M., Monier, C., Piwkowska, Z., Bal, T.,
  Frégnac, Y., Markram, H., and Destexhe, A.
\newblock Minimal {Hodgkin}–{Huxley} type models for different classes of
  cortical and thalamic neurons.
\newblock \emph{Biological Cybernetics}, 99\penalty0 (4):\penalty0 427--441,
  November 2008.

\bibitem[Prinz et~al.(2003)Prinz, Billimoria, and Marder]{prinz2003alternative}
Prinz, A.~A., Billimoria, C.~P., and Marder, E.
\newblock Alternative to hand-tuning conductance-based models: construction and
  analysis of databases of model neurons.
\newblock \emph{Journal of neurophysiology}, 2003.

\bibitem[Prinz et~al.(2004)Prinz, Bucher, and Marder]{prinz_similar_2004}
Prinz, A.~A., Bucher, D., and Marder, E.
\newblock Similar network activity from disparate circuit parameters.
\newblock \emph{Nature Neuroscience}, 7\penalty0 (12):\penalty0 1345--1352,
  December 2004.
\newblock Number: 12 Publisher: Nature Publishing Group.

\bibitem[Radev et~al.(2020)Radev, Mertens, Voss, Ardizzone, and
  Köthe]{radev_bayesflow_2020}
Radev, S.~T., Mertens, U.~K., Voss, A., Ardizzone, L., and Köthe, U.
\newblock Bayesflow: Learning complex stochastic models with invertible neural
  networks.
\newblock \emph{IEEE Transactions on Neural Networks and Learning Systems},
  2020.

\bibitem[Ramos et~al.(2019)Ramos, Possas, and Fox]{ramos2019bayessim}
Ramos, F., Possas, R.~C., and Fox, D.
\newblock Bayessim: adaptive domain randomization via probabilistic inference
  for robotics simulators, 2019.

\bibitem[Rauber et~al.(2017)Rauber, Brendel, and Bethge]{rauber2017foolbox}
Rauber, J., Brendel, W., and Bethge, M.
\newblock Foolbox: A python toolbox to benchmark the robustness of machine
  learning models.
\newblock In \emph{Reliable Machine Learning in the Wild Workshop, 34th
  International Conference on Machine Learning}, 2017.

\bibitem[Ribeiro et~al.(2022)Ribeiro, Zachariah, and
  Sch{\"o}n]{ribeiro2022surprises}
Ribeiro, A.~H., Zachariah, D., and Sch{\"o}n, T.~B.
\newblock Surprises in adversarially-trained linear regression.
\newblock \emph{arXiv preprint arXiv:2205.12695}, 2022.

\bibitem[Rozet et~al.(2021)]{rozet2021arbitrary}
Rozet, F. et~al.
\newblock Arbitrary marginal neural ratio estimation for likelihood-free
  inference.
\newblock \emph{arXiv preprint arXiv:2110.00449}, 2021.

\bibitem[Schmitt et~al.(2022)Schmitt, Bürkner, Köthe, and
  Radev]{schmitt_detecting_2022}
Schmitt, M., Bürkner, P.-C., Köthe, U., and Radev, S.~T.
\newblock Detecting {Model} {Misspecification} in {Amortized} {Bayesian}
  {Inference} with {Neural} {Networks}, November 2022.
\newblock arXiv:2112.08866 [cs, stat].

\bibitem[Shen et~al.(2019)Shen, Peng, Zhang, and Fan]{shen_defending_2019}
Shen, C., Peng, Y., Zhang, G., and Fan, J.
\newblock Defending against adversarial attacks by suppressing the largest
  eigenvalue of fisher information matrix.
\newblock \emph{arXiv preprint arXiv:1909.06137}, 2019.

\bibitem[Shen et~al.(2023)Shen, Ghosh, Sattigeri, Das, Bu, and
  Wornell]{shen-etal-2023-reliable}
Shen, M., Ghosh, S., Sattigeri, P., Das, S., Bu, Y., and Wornell, G.
\newblock Reliable gradient-free and likelihood-free prompt tuning.
\newblock In \emph{Findings of the Association for Computational Linguistics:
  EACL 2023}, pp.\  2416--2429, Dubrovnik, Croatia, May 2023. Association for
  Computational Linguistics.

\bibitem[Shu et~al.(2018)Shu, Bui, Zhao, Kochenderfer, and
  Ermon]{shu_amortized_2018}
Shu, R., Bui, H.~H., Zhao, S., Kochenderfer, M.~J., and Ermon, S.
\newblock Amortized {Inference} {Regularization}.
\newblock In \emph{Advances in {Neural} {Information} {Processing} {Systems}},
  volume~31. Curran Associates, Inc., 2018.

\bibitem[Sprungk(2020)]{sprungk_local_2020}
Sprungk, B.
\newblock On the {Local} {Lipschitz} {Stability} of {Bayesian} {Inverse}
  {Problems}.
\newblock \emph{Inverse Problems}, 36\penalty0 (5):\penalty0 055015, May 2020.
\newblock arXiv:1906.07120 [cs, math, stat].

\bibitem[Szegedy et~al.(2014)Szegedy, Zaremba, Sutskever, Bruna, Erhan,
  Goodfellow, and Fergus]{szegedy_intriguing_2014}
Szegedy, C., Zaremba, W., Sutskever, I., Bruna, J., Erhan, D., Goodfellow, I.,
  and Fergus, R.
\newblock Intriguing properties of neural networks.
\newblock In \emph{International Conference on Learning Representations}, 2014.

\bibitem[Tejero-Cantero et~al.(2020)Tejero-Cantero, Boelts, Deistler,
  Lueckmann, Durkan, Gonçalves, Greenberg, and Macke]{Tejero-Cantero2020}
Tejero-Cantero, A., Boelts, J., Deistler, M., Lueckmann, J.-M., Durkan, C.,
  Gonçalves, P.~J., Greenberg, D.~S., and Macke, J.~H.
\newblock sbi: A toolkit for simulation-based inference.
\newblock \emph{Journal of Open Source Software}, 5\penalty0 (52):\penalty0
  2505, 2020.

\bibitem[Trivedi \& Wang(2020)Trivedi and Wang]{trivedi2020expected}
Trivedi, S. and Wang, J.
\newblock The expected jacobian outerproduct: Theory and empirics, 2020.

\bibitem[Tsipras et~al.(2019)Tsipras, Santurkar, Engstrom, Turner, and
  Madry]{tsipras_robustness_2019}
Tsipras, D., Santurkar, S., Engstrom, L., Turner, A., and Madry, A.
\newblock Robustness may be at odds with accuracy.
\newblock In \emph{International Conference on Learning Representations}, 2019.

\bibitem[von Krause et~al.(2022)von Krause, Radev, and Voss]{von2022mental}
von Krause, M., Radev, S.~T., and Voss, A.
\newblock Mental speed is high until age 60 as revealed by analysis of over a
  million participants.
\newblock \emph{Nature human behaviour}, 6\penalty0 (5):\penalty0 700--708,
  2022.

\bibitem[Vovk(1990)]{vovk_aggregating_1990}
Vovk, V.~G.
\newblock Aggregating strategies.
\newblock In \emph{Proceedings of the third annual workshop on {Computational}
  learning theory}, {COLT} '90, pp.\  371--386, San Francisco, CA, USA, July
  1990. Morgan Kaufmann Publishers Inc.
\newblock ISBN 978-1-55860-146-8.

\bibitem[Ward et~al.(2022)Ward, Cannon, Beaumont, Fasiolo, and
  Schmon]{ward_robust_2022}
Ward, D., Cannon, P., Beaumont, M., Fasiolo, M., and Schmon, S.~M.
\newblock Robust neural posterior estimation and statistical model criticism.
\newblock In Oh, A.~H., Agarwal, A., Belgrave, D., and Cho, K. (eds.),
  \emph{Advances in Neural Information Processing Systems}, 2022.

\bibitem[Willetts et~al.(2021)Willetts, Camuto, Rainforth, Roberts, and
  Holmes]{willetts_improving_2021}
Willetts, M.~J., Camuto, A., Rainforth, T., Roberts, S., and Holmes, C.~C.
\newblock Improving {\{}vae{\}}s' robustness to adversarial attack.
\newblock In \emph{International Conference on Learning Representations}, 2021.

\bibitem[Yadan(2019)]{Yadan2019Hydra}
Yadan, O.
\newblock Hydra - a framework for elegantly configuring complex applications.
\newblock Github, 2019.

\bibitem[Zhang et~al.(2019)Zhang, Yu, Jiao, Xing, El~Ghaoui, and
  Jordan]{zhang_theoretically_2019}
Zhang, H., Yu, Y., Jiao, J., Xing, E., El~Ghaoui, L., and Jordan, M.
\newblock Theoretically principled trade-off between robustness and accuracy.
\newblock In \emph{International conference on machine learning}, pp.\
  7472--7482. PMLR, 2019.

\bibitem[Zhao et~al.(2019)Zhao, Fletcher, Yu, Peng, Zhang, and
  Shen]{zhao_adversarial_2019}
Zhao, C., Fletcher, P.~T., Yu, M., Peng, Y., Zhang, G., and Shen, C.
\newblock The adversarial attack and detection under the fisher information
  metric.
\newblock In \emph{Proceedings of the {AAAI} {Conference} on {Artificial}
  {Intelligence}}, volume~33, pp.\  5869--5876, 2019.
\newblock Issue: 01.

\end{thebibliography}
\bibliographystyle{icml2023}

%%%%%%%%%%%%%%%%%%%%%%%%%%%%%%%%%%%%%%%%%%%%%%%%%%%%%%%%%%%%%%%%%%%%%%%%%%%%%%%
%%%%%%%%%%%%%%%%%%%%%%%%%%%%%%%%%%%%%%%%%%%%%%%%%%%%%%%%%%%%%%%%%%%%%%%%%%%%%%%
% APPENDIX
%%%%%%%%%%%%%%%%%%%%%%%%%%%%%%%%%%%%%%%%%%%%%%%%%%%%%%%%%%%%%%%%%%%%%%%%%%%%%%%
%%%%%%%%%%%%%%%%%%%%%%%%%%%%%%%%%%%%%%%%%%%%%%%%%%%%%%%%%%%%%%%%%%%%%%%%%%%%%%%
\newpage
\appendix
\setcounter{figure}{0}
\renewcommand{\thefigure}{A\arabic{figure}}
\setcounter{section}{0}
\renewcommand{\thesection}{A\arabic{section}}

\onecolumn
\section*{\LARGE Appendix}

\section{Further experimental details}

\subsection{Training procedure}
\label{sec:appendix_training_procedure}
All methods and evaluations were conducted using PyTorch \citep{paszke_pytorch_2019}.
 %These parameters were then reparameterized to adhere to constraints. 
 We used Pyro \citep{bingham_pyro_2019} implementations of masked autoregressive flow (MAF) or rational linear spline flow (NSF) \citep{papamakarios_masked_2017,dolatabadi2020invertible}. We employed three transforms, each of which was parameterized using a two-layered ReLU multi-layer perceptron (MLP) with 100 hidden units. For higher dimensional tasks such as Hodgkin Huxley, VAE, and Spatial SIR, we used a ReLU MLP embedding network with 400, 200, and 100 hidden units and outputting a 50 dim vector.
 
The diagonal Gaussian model used two hidden layers, a hidden layer of size 100, and ReLU activations. The Multivariate Gaussian and mixture density networks used a hidden layer of size 200. 
 
 We trained each model with the Adam optimizer with a learning rate of $10^{-3}$, a batch size of $512$, and a maximum of 300 epochs. If training failed, the learning rate was reduced to $10^{-4}$ or $10^{-5}$. To prevent overfitting, we used early stopping based on a validation loss evaluated on $512$ hold-out samples. Each model was trained on the same set of either $10^3, 10^4$, or $10^5$ simulations.

For the adversarial attacks, we performed 200 projected gradient descent steps and estimated the $D_{KL}$ with a Monte Carlo average of 5 samples at each step (for the Gaussian models we used the analytical solution).  Additionally, each adversarial example was clamped to the minimum and maximum values within the test set to avoid generating samples outside of the support of $p(\vx)$. After the optimization was finished, for the evaluation of the attack, the adversarial objective was evaluated with a Monte Carlo budget of 256 samples. We note that, as the scale of data varies strongly between simulators, all tolerance levels $\epsilon$ were normalized by the average of the standard deviation of prior predictives. Table \ref{tab:abs_tol} shows the resulting tolerance levels.

For the results shown in Fig.~\ref{fig:defenses_attacked}, we used a different value for the regularization strength $\beta$ for each task, as the parameter is coupled to the magnitude of the perturbation (which is different for each task). The particular value of $\beta$ was hand-picked based on initial experiments. The values were: $\beta=0.001$ for Gaussian linear, $\beta=0.1$ for SIR, $\beta=0.01$ for Lotka Volterra, $\beta=100$ for Hodgkin Huxley, $\beta=0.01$ for VAE and $\beta=0.1$ for Spatial SIR. Sweeps for $\beta$ for each task are shown in Fig.~\ref{fig:fim_hyperparameters}. We used $5$ Monte Carlo samples per iteration and a momentum of $\gamma=0.85$ for each benchmark task for the FIM regularizer. We used a MAF for all benchmark results in the main paper and evaluated different density estimators in Figs.~\ref{fig:fig3_appendix} and \ref{fig:coverages_per_density estimator}.

For the pyloric network task, we employed a MAF with three transforms, each parameterized by a 3-layer neural network with 200 hidden neurons. We also utilized an embedding network composed of three 1D convolutional neural networks, each with three convolutional layers that produce six, nine, and twelve output channels. These networks were applied to the voltage trace of each neuron. The results were then summarized by a 3-layer feed-forward neural network, and reduced to a 100-dimensional feature vector. We trained the model using 750,000 pre-selected simulations and evaluated its convergence on a validation set of 4096 additional datapoints. The evaluation was conducted on 10,000 separate simulations. For the FIM regularized model, we used $\beta = 100$. Further, we set the number of Monte Carlo samples within the exponential moving average to one; the momentum remained $\gamma=0.85$.

% TODO: hyperparameters for defenses: how many Monte Carlo samples for FIM and TRADES on each benchmark task and on pyloric. Also, what value are we using for gamma (momentum)?

\subsection{Benchmark tasks}
\label{sec:appendix_benchmark}
We used the following benchmark tasks, which produce relatively complex and high-dimensional data. We used these tasks instead of established benchmark tasks \citep{lueckmann_benchmarking_2021} because our tasks are chosen to have more high-dimensional data and, thus, might offer more flexibility for adversarial attacks.

 \paragraph{Gaussian Linear:} A simple diagonal linear mapping (entries sampled from a standard Gaussian) of a ten-dimensional vector subject to isotropic Gaussian noise:
 $$ p(\vx, \vtheta) = \mathcal{N}(\vx;\mathbf{A}\vtheta, \sigma^2 \mathbf{I} ) \mathcal{N}(\vtheta, \mathbf{0}, \mathbf{I})$$
 with $\sigma = 0.1$. As a result, the posterior is also Gaussian and analytically tractable.

\paragraph{Lotka Volterra:} An ecological predator-prey model with four parameters. It is given by the solution of the following differential equation:
$$ \frac{dx}{dt} = \alpha x - \beta xy \quad \frac{dy}{dt} = \delta xy -\gamma y  $$
with  $\alpha$ representing the growth rate of prey, $\beta$ the death rate of prey, $\delta$ the hunting efficiency of the predator and $\gamma$ the death rate of the predator.
The observed data are the predator and prey population densities at 50 equally spaced time points. We added normally distributed noise with $\sigma=0.05$. The prior for the parameters is Gaussian with $\mu = 0.$ and $\sigma=0.5$, transformed by a sigmoid function in the simulator to be positive and bounded. The resulting 4-dimensional posterior is highly correlated.
\paragraph{VAE:} The decoder $g_\psi(x)$ of a Variational Autoencoder (VAE) was used as a generative model for handwritten digits \citep{kingma_auto-encoding_2022}. The prior is a three-dimensional standard Gaussian. Images generated by the decoder were used as observed data ($28 \times 28$ dimensional), with Gaussian observation noise with $\sigma = 0.05$:
$$ p(\vx, \vtheta) = \mathcal{N}(\vx; g_\psi(\vx), \sigma^2 \mathbf{I}) \mathcal{N}(\vtheta; \mathbf{0}, \mathbf{I})$$
 Due to the training procedure of the VAE, the posterior should be almost Gaussian. 

\paragraph{Hodgkin Huxley:}  A neuroscience model that describes how action potentials in neurons are initiated and propagated. It is implemented based on \citet{pospischil_minimal_2008} and taken from \citet{Tejero-Cantero2020}. We use a uniform prior constrained to biologically reasonable values. The observed data is the membrane voltage at 200 equally spaced time points, to which we added normally distributed noise with $\sigma=0.1$, leading to a 7-dimensional posterior distribution.
\paragraph{SIR:} A epidemiological model with two free parameters: The rate of recovery for infected individuals $\gamma$ and the rate of new infections $\beta$ for a population of $N=5$. The solution satisfies the following differential equation
$$ \frac{dS}{dt} = - \beta \frac{S \cdot I}{N} \quad \frac{dI}{dt} = \beta \frac{S \cdot I}{N}  \gamma  - \gamma I  \quad \frac{dI}{dt} = \gamma I. $$
The observed data correspond to the number of infections at 50 equally spaced time points. We added log-normal observation noise with $\sigma=0.2$. The prior was a Gaussian with $\sigma=2.$ transformed by a sigmoid function,  resulting in a complex 2-dimensional posterior.
\paragraph{Spatial SIR:} An epidemiological model with a spatial dimension, similar to \citet{hermans_trust_2022}. The model is initialized with three infections at random locations on the grid. The infection then propagates to neighboring grid cells with probability $\beta$ per time step. Infected people recover with probability $\gamma$ at each time step. We observed a $30 \times 30$ grid of infected/non-infected regions, subject to a Beta noise model modeling the probability of being infected after a test. A logNormal prior with $\sigma=0.5$ was used.

\subsection{Metrics}
\label{sec:appendix_metrics}
 The expected coverage was evaluated as proposed in \citet{cannon_investigating_2022, hermans_trust_2022}. Given the $100(1-\alpha)\%$ highest posterior density region of the posterior estimate $\text{HPR}_{1-\alpha}$, we target to estimate
 $ \mathbb{E}_{p(\vtheta,\vx)} \left[ \mathds{1} \{ \vtheta \in \text{HPR}_{1-\alpha}\} \right]$, where $\mathds{1}$ is the indicator function. If the model is well-calibrated, the empirical coverage should match the nominal coverage $1-\alpha$.
 
 As in \citet{cannon_investigating_2022}, we evaluate the coverage given (adversarially) perturbed data. Thus, the expected coverage becomes:
 $$   \mathbb{E}_{p(\tilde{\vx})} \left[\mathbb{E}_{p(\vtheta|\tilde{\vx})} \left[ \mathds{1}\{ \vtheta \in \text{HPR}_{1-\alpha}\} \right] \right]$$
%Since the inner expectation evaluates to $1-\alpha$, we can compute this quantity for any $\tilde{\vx} \sim p(\tilde{\vx})$. 
We use a Monte Carlo approximation to obtain $\text{HPR}_{1-\alpha}$ as described by \citet{rozet2021arbitrary, deistler2022truncated} to efficiently estimated this quantity.

\begin{table}
\caption{\textbf{Relative and absolute tolerance levels for each task.} The values of $\epsilon$ are multiples of the average standard deviation of prior predictives.}
\label{tab:abs_tol}
\vskip 0.15in
\begin{center}
\begin{small}
\begin{sc}
\begin{tabular}{lrrrrrr}
\toprule
Task & $\epsilon = 0.1$ & $\epsilon = 0.2$ & $\epsilon = 0.3$ & $\epsilon = 0.5$ & $\epsilon = 1$ & $\epsilon = 2$ \\
\midrule
Gaussian Linear   &  0.11& 0.21& 0.32 & 0.53 & 1.05 & 2.13 \\
SIR               & 0.03 &0.06& 0.09 & 0.15 & 0.3 & 0.6\\
Lotka Volterra    & 0.02 &0.04& 0.06 & 0.1 & 0.19 & 0.39\\
Hodgkin Huxley    & 1.83 &3.65& 5.48 & 9.13 &  18.26& 36.54\\
VAE               & 0.02 &0.05& 0.07 & 0.12 & 0.25 & 0.5\\
Spatial SIR       & 0.02 &0.04& 0.07 & 0.11 & 0.22 & 0.45\\
\bottomrule
\end{tabular}
\end{sc}
\end{small}
\end{center}
\vskip -0.1in
\end{table}

\begin{figure}
    \centering
    \includegraphics[width=\textwidth]{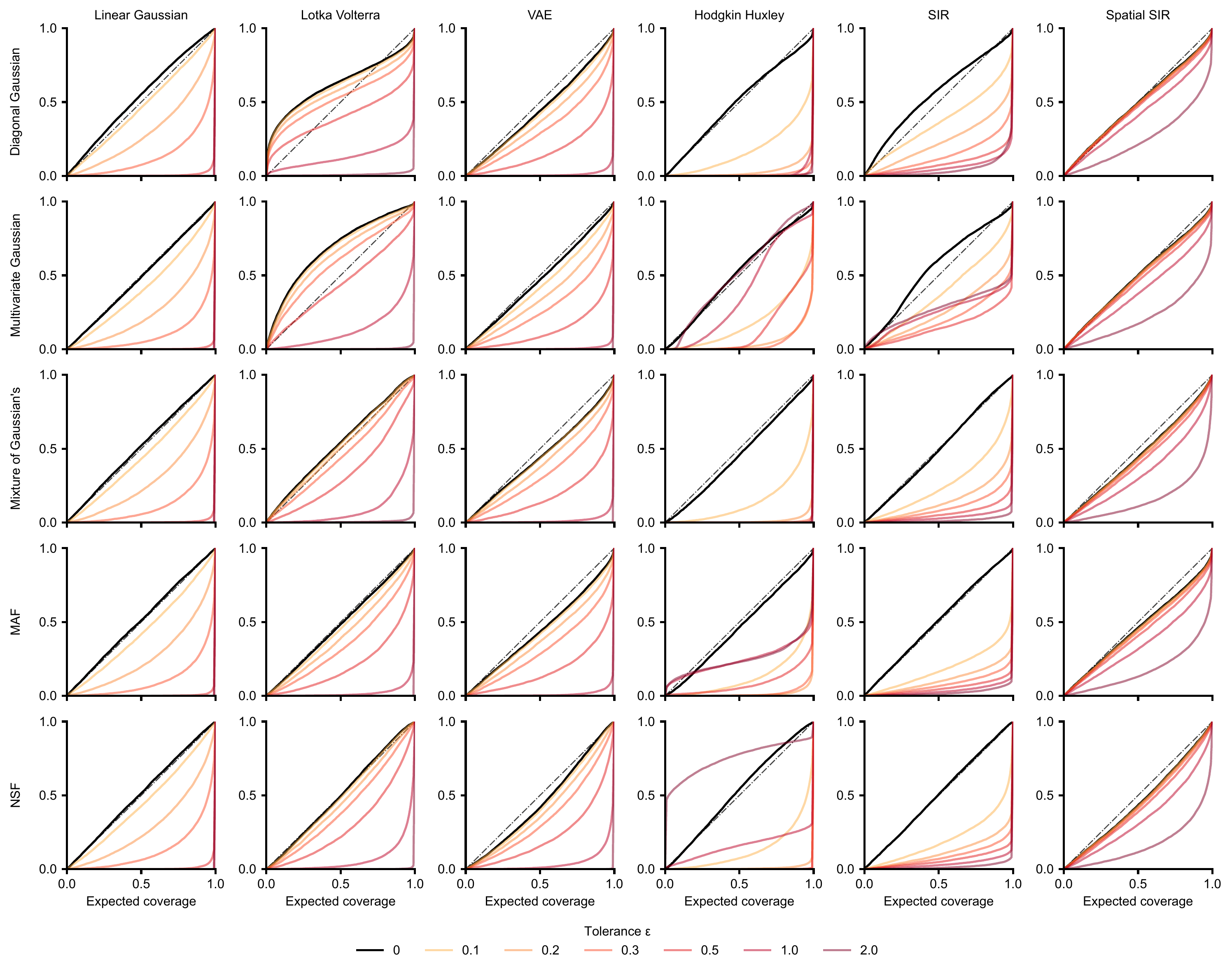}
    \caption{\textbf{Expected coverages for all density estimators.} Expected coverage metric for a specific density estimator trained with $10^5$ simulations on standard NPE loss. The performance on well-specified data in black; colors indicate performance on adversarially perturbed data at certain tolerance levels.}
    \label{fig:coverages_per_density estimator}
\end{figure}

\section{Adversarial training and TRADES}
\label{appendix:defense_methods}

We mainly compared against two well-known methods to defend against adversarial examples (additional defenses and results  in Sec.~\ref{sec:appendix_defenses}).

\subsection{Adversarial training}
\citet{madry_towards_2019} proposed that the loss should be modified such that, instead of minimizing the negative log-likelihood (NLL) given the clean observation, one minimizes the NLL given the worst possible observation $\tilde{x}$ within an $\epsilon$ ball around the observation (i.e. the observation that has the highest NLL within the $\epsilon$ ball). Formally, this objective can be written as
$$ \min_{\phi} \mathbb{E}_{p(\vx,\vtheta)} \left[ \max_{\tilde{\vx}\in \{\tilde{\vx} \ \mid \ || \tilde{\vx} - \vx|| \leq \epsilon \}} - \log q_\phi (\vtheta|\tilde{\vx} ) \right].$$
This scheme encourages the trained neural network to have a low NLL for any $\tilde{\vx}$ within the $\epsilon$-ball and thus encourages robustness to any adversarial perturbation.

In order to find the $\tilde{\vx}$ with the highest NLL within the $\epsilon$-ball, one commonly generates an adversarial example \textit{during training} e.g., with projected gradient descent \citep{madry_towards_2019}. Considering the adversarial perturbation as a distribution $p(\tilde{\vx}|\vx, \vtheta)$ gives:
%Relaxations is proposed by instead generating an adversarial example, approximately solving the inner maximization problem. Most adversarial attacks are inherently stochastic either due to random initialization or objective function, thus specify a conditional density $p(\tilde{\vx}|\vx, \vtheta)$ (or $p(\tilde{\vx}|\vx)$ depending on the attack). 
%Equivalently the loss can thus be written as
\begin{align*}  
\mathcal{L}(\phi) &= \mathbb{E}_{p(\vx,\vtheta)}  \left[ \mathbb{E}_{p(\tilde{\vx}|\vx,\vtheta)} \left[  - \log q_\phi (\vtheta|\tilde{\vx} )  \right]\right] \\&= \mathbb{E}_{p(\tilde{\vx},\vtheta)} \left[  - \log q_\phi (\vtheta|\tilde{\vx} )  \right] 
\end{align*}
% Technically p(\tilde{x}|x,\vtheta) does also depend on \phi, but 
This reveals that by modifying the training scheme, $q_\phi$ no longer converges to the true posterior but to a different posterior distribution given by
$$ \tilde{p}(\vtheta|\tilde{\vx}) \propto \tilde{p}(\tilde{\vx}|\vtheta)p(\vtheta) = \int p(\tilde{\vx}|\vx,\vtheta)p(\vx|\vtheta) d\vx p(\vtheta)$$
i.e., the posterior distribution given the likelihood of observing the adversarially perturbed $\tilde{\vx}$ given $\vtheta$. Therefore, adversarial training can be interpreted as a regularization scheme where the data is perturbed by the adversarial perturbation (instead of a random perturbation as in \citet{shu_amortized_2018}).

In our experiments, we use an $\ell_2$ projected gradient descent attack with 20 iterations during training. 
%We treat the tolerance level $\epsilon$ as a hyperparameter, which we evaluate for $0.1$, $0.5$, and $1.0$ during training. 
After initial experiments, we hand-picked $\eps = 0.1$ for the Gaussian linear task and $\eps=1.0$ for the other benchmark tasks during training.

\subsection{TRADES}
A second method for adversarial robustness was proposed by \citet{zhang_theoretically_2019}. Their proposed loss function balances the trade-off between performance on clean and adversarially perturbed observations, controlled through a hyperparameter $\beta$. The resulting surrogate loss adds a Kullback-Leibler divergence regularizer in the form of
\begin{align*}
\mathcal{L}(\phi)= \mathbb{E}_{p(\tilde{\vx},\vx,\vtheta)} \left[ - \log q_\phi (\vtheta|\vx ) + \beta D_{KL}(q_\phi(\vtheta|\vx)||q_\phi(\vtheta|\tilde{\vx}) \right].
\end{align*}
This ensures that the posterior estimate is smooth (as measured by the KL divergence). However, this approach requires the ability to evaluate the KL divergence, which, for normalizing flows, can only be approximated through Monte Carlo techniques. 

In general, the strength of regularization is determined by both the magnitude of the adversarial example, $\epsilon$, and the hyperparameter $\beta$. Initial experiments showed that $\epsilon$ had a greater effect than $\beta$, so we only varied $\epsilon$. To estimate the $D_{KL}$ during training, we use a single Monte Carlo sample.
%to estimate the KL divergence term during training. 
We fixed $\beta=0.1$ and hand-picked, based on initial experiments, $\eps=0.1$ for the linear Gaussian and VAE, $\eps=0.5$ for Hodgkin Huxley, Lotka Volterra, and Spatial SIR.
The optimization to obtain $\tilde{\vx}$ was run with an $\ell_2$ projected gradient descent attack with 20 iterations.

\section{Tradeoff between posterior approximation and robustness}
\label{sec:appendix_tradeoff}

Adversarially robust models typically sacrifice accuracy on clean data in order to achieve robustness. The errors on clean and perturbed data have even been suggested to be fundamentally
at odds \citep{zhang_theoretically_2019, tsipras_robustness_2019} and are subject to the required strength of adversarial robustness \citep{min_curious_2020}, even in the infinite data limit.

Regularizing with the Fisher Information Matrix (FIM) creates a similar trade-off between accuracy on clean and perturbed data. The standard loss of neural posterior estimation minimizes 
$$\mathcal{L}(\phi) =\mathbb{E}_{p(\vx)} \left[ D_{KL}(p(\vtheta|\vx)||q_\phi(\vtheta|\vx))\right] = \mathbb{E}_{p(\vx,\vtheta)}\left[ -\log q_\phi (\vtheta|\vx)\right]$$ whereas the FIM regularizer minimizes 
$$\Omega (\phi) = \mathbb{E}_{p(\vx,\tilde{\vx})} \left[ D_{KL}(q_\phi(\vtheta|\vx)||q_\phi(\vtheta|\tilde{\vx}))\right].$$

$\Omega(\phi)$ is minimized globally if $q_\phi(\vtheta|\vx) = q_\phi(\vtheta|\tilde{\vx})$ for all $\vx, \tilde{\vx} \sim p(\vx)$. This suggests that the optimal $q_\phi(\vtheta|\tilde{\vx})$ is independent of $\vx$ and, thus, indeed at odds with approximating the posterior distribution given clean data. 

The strength of this trade-off is determined by the value of the hyperparameter $\beta$, and the effect on the posterior fit is demonstrated in Fig.~\ref{fig:defenses_attacked}C, were we plot the trade-off between accuracy on clean data (evaluated as average log-likelihood) and the robustness to adversarial perturbations (measured as $D_{KL}$ between clean and perturbed posteriors). The plotted values of $\beta$ are Pareto-optimal solutions approximately solving the multi-objective optimization problem: 

$$ \min_{\beta} \left[ \mathcal{L}(\phi_\beta^*) , \Omega(\phi_\beta^*)\right] \text{ with } \phi_\beta^* = \argmin_{\phi_\beta} \mathcal{L}(\phi_\beta) + \beta \mathbb{E}_{p(\vx)}\left[ tr(\mathcal{I}_\vx) \right]$$

It is clear that a large value of $\beta$ heavily regularizes $\Omega$, pushing it towards zero, which results in the inference model ignoring the data and increasing $\mathcal{L}$. Thus, as $\beta$ grows large, $q_\phi(\vtheta|\vx)$ approaches $p(\vtheta)$, i.e. the prior distribution, as this is the best estimate according to $\mathcal{L}$ which is independent of $\vx$. This is in correspondence with approaches for robust generalized Bayesian inference with, e.g., $\alpha$-posteriors \citep{grunwald_inconsistency_2017, vovk_aggregating_1990, medina_robustness_2021}.

However, for smaller values of $\beta$, there is a plateau where the accuracy of clean data is almost constant, but the levels of robustness vary significantly. This suggests that multiple inference models exist that have similar approximation errors but differ in their robustness to adversarial examples. The regularizer in this region can effectively induce robustness without sacrificing much accuracy.

It is worth noting that at a certain value of $\beta$, the approximation error increases significantly while the robustness decreases only gradually. As previously discussed in the main paper, $\beta$ is closely related to the magnitude of the adversarial perturbation $\tilde{\vx}$ i.e. the tolerance level $\epsilon$. The \emph{true} posterior might not be robust to such large-scale perturbations, making it an invalid solution subject to robustness constraints. 

% \begin{figure}
%     \centering
%     \includegraphics[width=\textwidth]{figures/fig_tradeoff_appendix.pdf}
%     \caption{Shows the tradeoff between the objective function (i.e. posterior approximation) and robustness for different choices of regularization strengths $\beta$ of the FIM regularizer.}
%     \label{fig:tradeoff}
% \end{figure}

\section{Additional benchmark results}

Here, we present additional results obtained on the benchmark tasks.

\subsection{Runtime}
\label{sec:runtime}
\begin{figure}
    \centering
    \includegraphics[width=\textwidth]{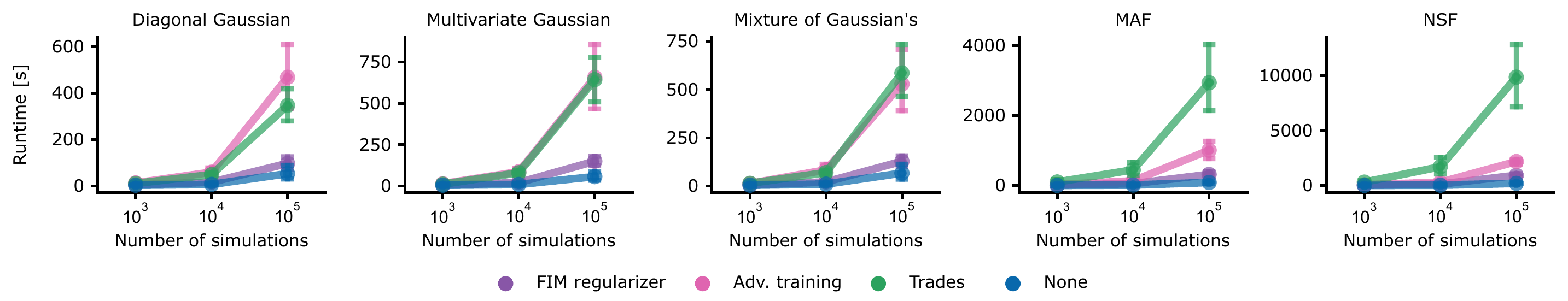}
    \caption{\textbf{Runtime.} Average runtime in seconds for each defense, calculated across all benchmark tasks and various hyperparameters per density estimator. For the MAF estimator with 5 Monte Carlo samples, the average timings for $100k$ training simulations are: 60 seconds for NPE, 250 seconds for FIM, 770 seconds for adv.~training, and 2790 seconds for TRADES.}
    \label{fig:runtime}
\end{figure}

In Figure \ref{fig:runtime}, we show the average runtime of the benchmark tasks. It can be observed that FIM regularization has a slightly higher cost compared to standard NPE, whereas TRADES is substantially more expensive across various density estimators. This especially holds for normalizing flows where the $D_{KL}$ regularizer is estimated via Monte Carlo.

\subsection{Additional results on attacks}
\label{sec:attack_additional_results}

\begin{figure}
    \centering
    \includegraphics[width=\textwidth]{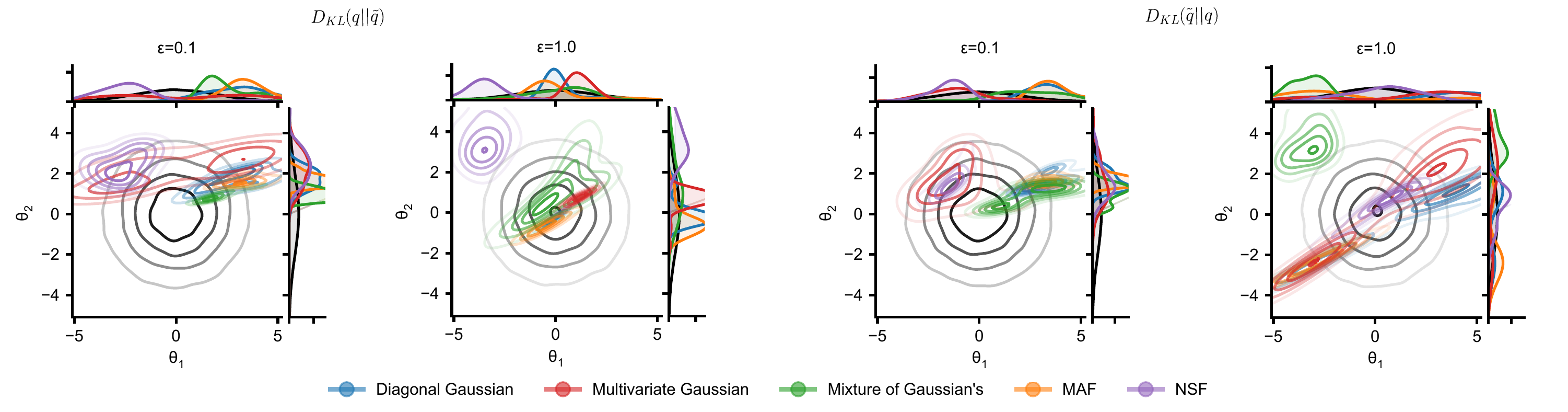}
    \caption{\textbf{Parameters which generated data susceptible to adversarial perturbations.} In black, we illustrate the prior over parameters $\vtheta$ for the SIR task. For each density estimator trained with $10^5$ simulations, we plot the distribution of $\vtheta$'s on which we found the 50 "strongest" adversarial examples using the L2PGDAttack maximizing $D_{KL}(q||\tilde{q})$ (left) or $D_{KL}(\tilde{q}||q)$ (right).}
    \label{fig:attack_thetas}
\end{figure}

\paragraph{Which observations are particularly vulnerable to adversarial attacks?} In our experiments, we generated a subset of parameters $\vtheta$, along with well-specified data points $\vx$, and adversarial examples found on these data points, $\tilde{\vx}$. 
%Notably, the attacks do not require knowledge of $\vtheta$, making them statistically independent. This allows us to directly identify the parameters that generate data that is sensitive to adversarial perturbations. 
We next asked which observations are particularly vulnerable to adversarial attacks and what regions these correspond to in parameter space. We selected the 10\% datapoints which had the highest $D_{KL}$ between posterior estimates given clean and perturbed data and visualized the distribution of their corresponding ground truth parameters (Fig.~\ref{fig:attack_thetas}). Attacks on different density estimators have higher efficacy on different sets of parameters, with some similarities (especially for similar density estimators such as Gaussian and Multivariate Gaussian). This indicates that the attacks not only leverage worst-case misspecification but also attack the particular neural network. Notably, parameters that generate data that is vulnerable to adversarial attacks are not necessarily found in the tails of the prior (where training data is scarce), but also in regions where many training data points are available. This observation could imply either vulnerable areas in the specific neural networks or/and susceptible regions within the generative model.

\begin{figure}
    \centering
 \includegraphics[width=\textwidth]{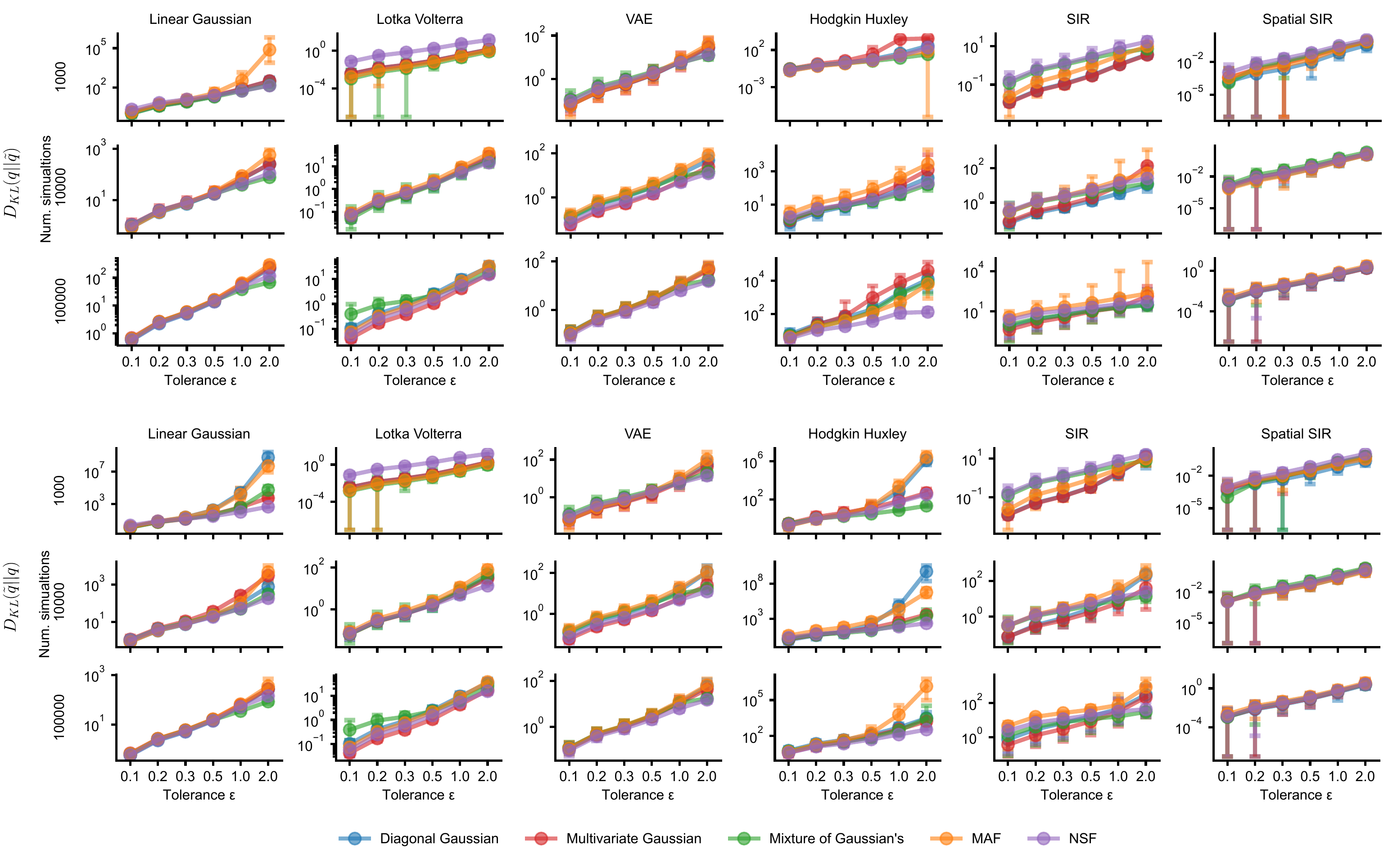}
    \caption{\textbf{Robustness for different density estimators.} Each density estimator is trained with NPE loss with a different number of simulations, attacked by an $\ell_2$ projected gradient descent attack trying to maximize $D_{KL}(q||\tilde{q})$ (top) or $D_{KL}(\tilde{q}||q)$ (bottom).}
    \label{fig:fig3_appendix}
\end{figure}

\paragraph{Are more complex density estimators less robust?} We evaluated the $D_{KL}$ (forward and backward) between posterior estimates given clean and perturbed data for several conditional density estimators. Our results show similar adversarial robustness across all tested density estimators (Fig.~\ref{fig:fig3_appendix}). 
%Yet, every inference model is attacked by the same adversary, which may be to weak for the larger models. 
We note that attacking simple models might appear to be more vulnerable because the adversarial objective can be computed in closed form, whereas complex models require Monte Carlo approximations.

We also computed the expected coverage for all density estimators (Fig.~\ref{fig:coverages_per_density estimator}). Again, the expected coverages suggest a similar level of adversarial robustness across different conditional density estimators. %Overall they behave quite similar. Note that not all density estimator are able to solve each task perfectly as they are limited in what type of distribution they can model, leading to imperfect calibration without on clean data. 

\paragraph{Does the adversarial objective matter?}
\label{different_adversarial_attacks}
We evaluated whether using the forward vs the backward $D_{KL}$ as the target for the adversarial attack influences the results. Despite minor differences, there is no clear advantage of divergence over the other. Adversaries with different objectives may find different adversarial examples that are more severe, as measured by their notion of "distance" between the distributions, as shown in Figure \ref{fig:fig3_appendix}. As the KL divergence is locally symmetric, these differences are only noticeable for larger tolerance levels $\epsilon$.

In addition, we evaluated an attack based on the Maximum-Mean discrepancy (MMD)
$$ \delta^* = \arg\max_\delta MMD^2(q_\phi(\theta|\vx) || q_\phi(\vtheta, \vx + \vdelta)) \text{ s.t. } ||\delta||\leq \epsilon .$$
We use the kernel MMD with an RBF kernel, estimated by a linear time Monte Carlo estimator as defined in \citet{gretton_optimal_2012} using ten samples. The MMD attack has a similar impact as $D_{KL}$ attacks, but it is significantly weaker for some tasks, such as Lotka Volterra and SIR. One potential explanation for this could be an unsuitable kernel selection. Specifically, if the length scale of the RBF kernel is too small, it is well-known that the gradients tend to vanish. This issue can be particularly noticed in the SIR task, which plateaus for larger tolerance levels (in contrast to KL-based attacks, which explode). We note, however, that the MMD attack could also be applied to implicit density estimators (such as VAEs or GANs)
%Our research primarily centers around explicit density estimators, which are models that can compute density. However, implicit models, which cannot directly evaluate density, are also employed by several methods. Although our attacks utilize the KL divergence, several other divergences could also be utilized. We anticipate that for small tolerance levels, each member of the f-divergence family will behave similarly since they are locally equivalent up to a scaling constant. A different family of divergences are integral probability metrics, for example, the Maximum-Mean discrepancy (MMD). One advantage of the MMD is that it can be estimated only using samples, hence is also tractable for implicit models.

Finally, we evaluated an attack that minimizes the log-likelihood of the true parameters:
$$ \delta^* = \arg\max_\delta -\log q_\phi(\vtheta_o|\vx_o + \vdelta) \text{ s.t. } ||\delta||\leq \epsilon . $$
Note that his attack requires access to one true parameter, which is only available for observations generated from the model and hence is not generally applicable. Furthermore, minimizing the likelihood of a single good parameter may not inevitably decrease the likelihood of all probable parameters. This attack strongly impacts the expected coverage since the attack objective is explicitly designed to avoid the true parameter and push it away from the region of the highest density (which is precisely the quantity measured by this metric).

\label{sec:appendix_other_divergences}
\begin{figure}
    \centering
    \includegraphics[width=\textwidth]{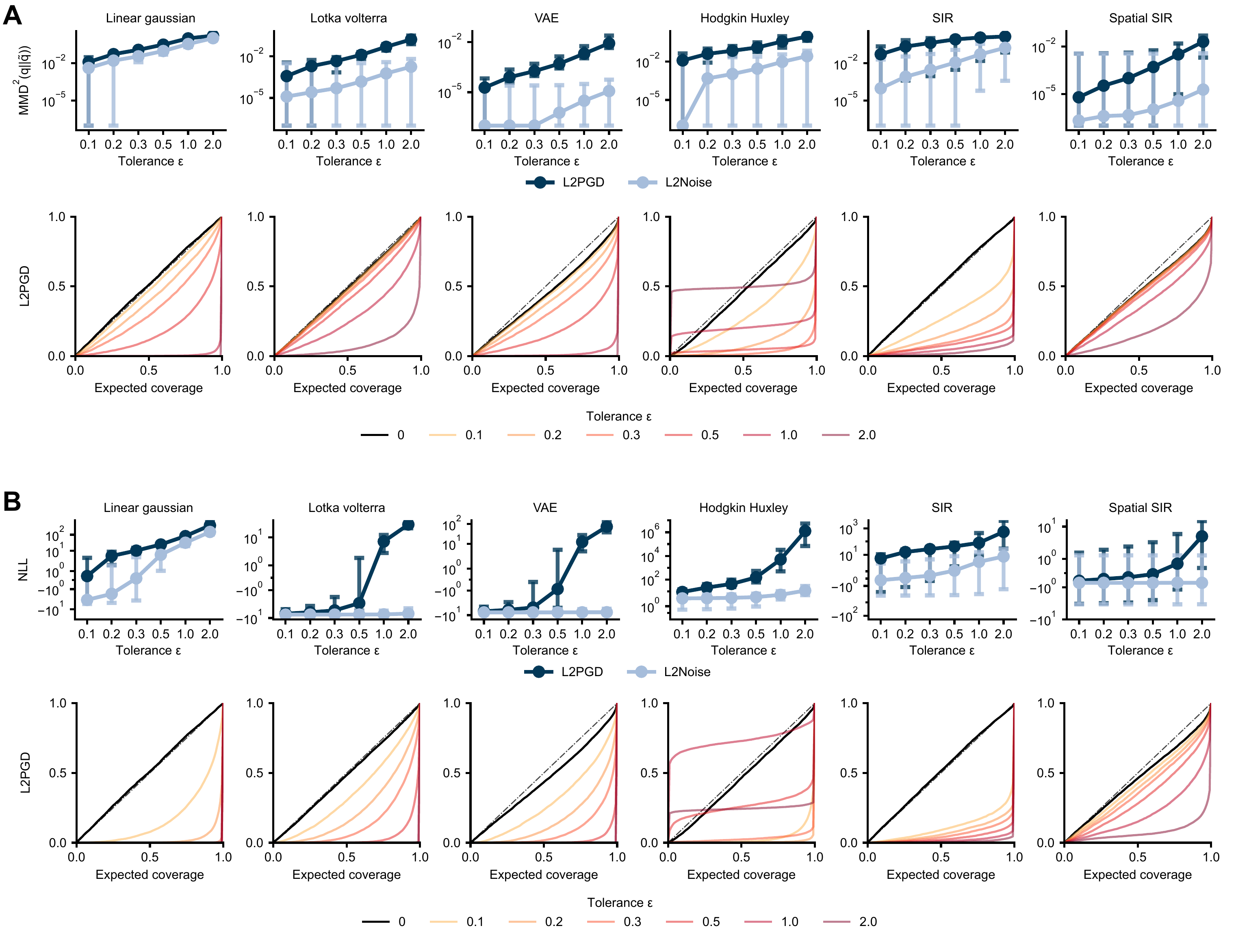}
    \caption{\textbf{Adversarial attack using different divergence.} Adversarial attacks using on each task using the MMD (\textbf{A}) and minimizing the true parameters log likelihood (\textbf{B}). On top, we show robustness $D_{KL}(q||\tilde{q})$ at the bottom; we show the expected coverage. The performance on well-specified data is shown in black; the colors indicate performance on adversarially perturbed data at certain tolerance levels.}
    \label{fig:mmd_mle_attack}
\end{figure}

% \begin{figure}
%     \centering
%     \includegraphics[width=\textwidth]{figures/expected_coverages_fKL_n100000.pdf}
%     \caption{Caption}
%     \label{fig:my_label}
% \end{figure}

\subsection{Additional results for the defenses}
\label{sec:appendix_defenses}

\paragraph{Robustness of ensembles and noise augmentation}

In addition to adversarial training and TRADES, we investigated two defense methods that were not originally developed as defenses against adversarial attacks: Ensembles and Noise Augmentation. 

Ensembles do not use a single density estimator but $K$ different ones. Assuming we trained all $K$ density estimators to estimate the posterior distribution on different initialization, thus falling into different local minima. Then an Ensemble Posterior is typically defined as
$$ q(\vtheta|\vx) = \sum_{k=1}^K \frac{1}{K} q_{\phi_k}( \vtheta | x)$$

We built an ensemble of $10$ masked autoregressive flows and evaluated its robustness to adversarial attacks on the benchmark tasks (Figure \ref{fig:appendix_defenses_adv_training_trades}). The ensemble has a similar robustness as standard NPE. 

Another defense, called `Noise Augmentation', adds random perturbations to the data during training (in contrast to adversarial training, which uses adversarial perturbations). We use random noise uniformly distributed on the $\ell_2$ ball with $\epsilon = 1.0$. Again, this defense only slightly (if at all) improved the robustness of NPE. %Adversarial training, in contrast, uses adversarial instead of random perturbations of similar or smaller magnitudes, with a much more noticeable impact on the posterior estimate and the adversarial robustness. Trades and FIM regularization follow this trend with an overall more substantial effect.

Overall, these results show that these defenses are not suitable to make amortized Bayesian inference robust to adversarial attacks.

\paragraph{Visualizing adversarial attacks with defenses.}

To visualize the effect on the inference of adversarial examples, we reproduced Figure \ref{fig:adversarial_examples} but with the robust inference models (with $\epsilon=1.0$). FIM regularization is an effective defense while maintaining good accuracy on clean data, as evidenced by reasonable predictive distributions (Fig.~\ref{fig:fig2_appendix_fisher}). Notably, both adversarial training and TRADES do not work as well (Fig.~\ref{fig:fig2_appendix_advtrain}, Fig.~\ref{fig:fig2_appendix_trades}).

Please note that in these figures, we do not necessarily utilize the identical observation or posterior. Instead, we present visual representations of examples that share the same "rank". By sorting all adversarial perturbations based on their adversarial objective, we display the outcomes that correspond to the same index as selected for Figure~\ref{fig:adversarial_examples}. Pyloric network examples were additionally constrained to be biophysically realistic.
% Especially TRADES might also suffer form stochasticity introduced by the Monte Carlo approximated KL divergence.

% \paragraph{Effect of the regularization strength}
% Figure \ref{fig:fim_hyperparameters} illustrates the effects of varying the regularization strength $\beta$ on the robustness of the inference model. The level of regularization needed varies depending on the task, with larger values typically required for data with a larger scale. Therefore, we used the highest regularizer for the Hodgkin-Huxley task.

% Unfortunately, expected coverage is not the best metric to evaluate the success of the defense model as the prior distribution is perfectly calibrated, but of course a trivial but most robust solution.

%%%%%%%%%%%%%%%%%%%%%%%%%%%%%%%%%%%%%%%%%%%%%%%%%%%%%%%%%%%%%%%%%%%%%%%%%%%%%%%
%%%%%%%%%%%%%%%%%%%%%%%%%%%%%%%%%%%%%%%%%%%%%%%%%%%%%%%%%%%%%%%%%%%%%%%%%%%%%%%

\begin{figure}
    \centering
    \includegraphics[width=\textwidth]{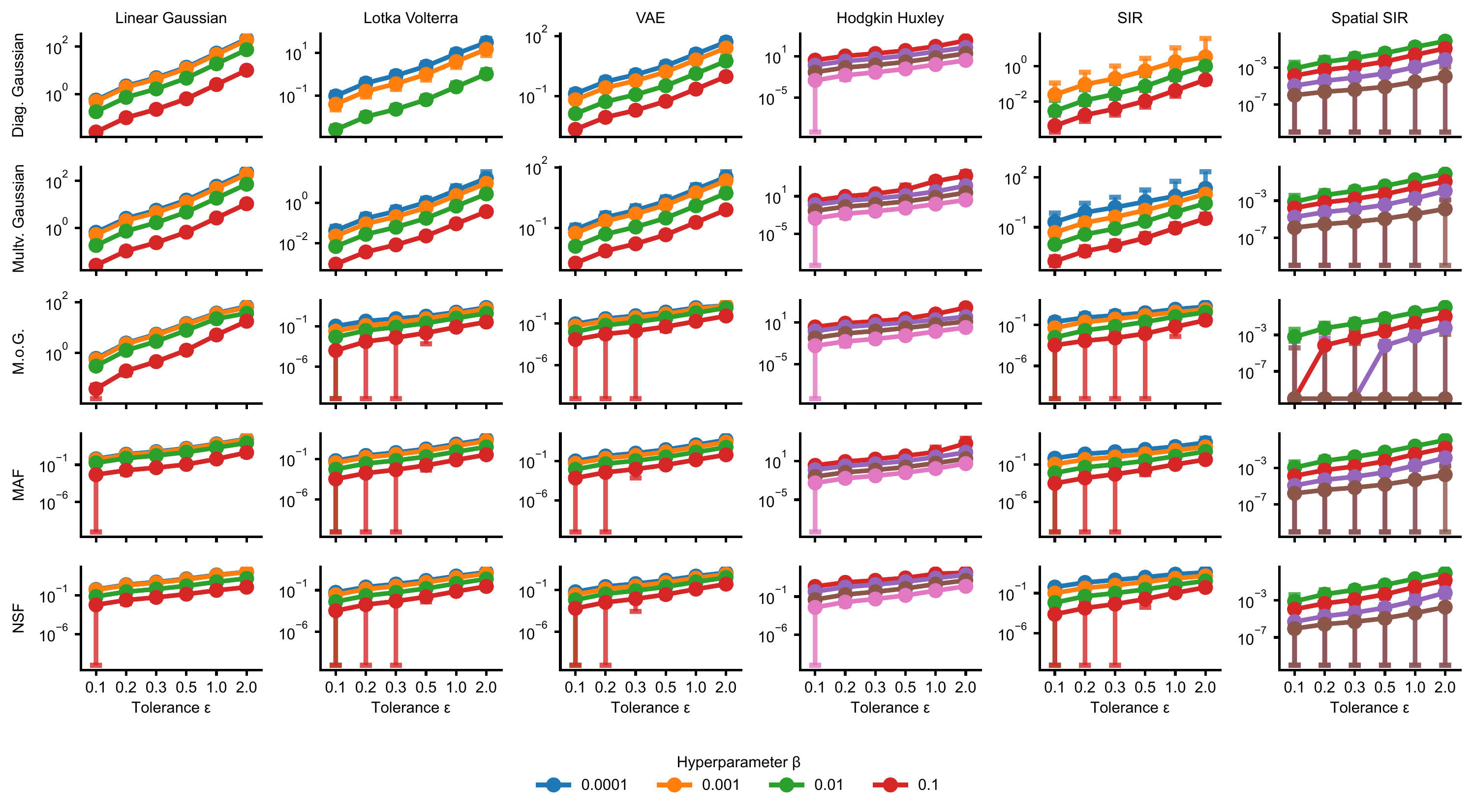}
    \caption{\textbf{Robustness of FIM.} Shows $D_{KL}(q||\tilde{q})$ for different density estimators (rows) and different values of regularization strength $\beta$ (colors).}
    \label{fig:fim_hyperparameters}
\end{figure}

\begin{figure}
    \centering
    \includegraphics[width=\textwidth]{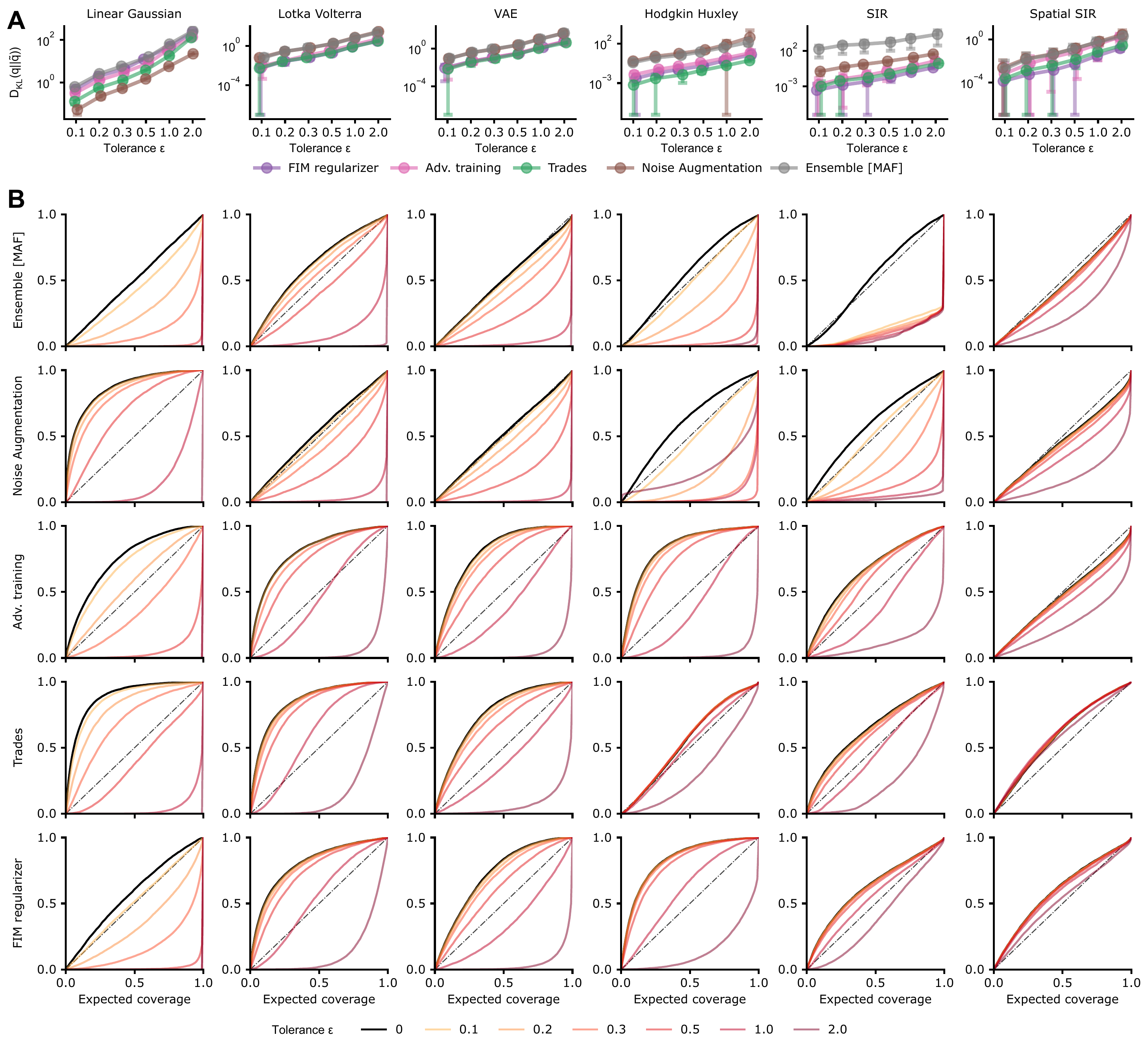}
    \caption{\textbf{Results for other defenses}. Robustness $D_{KL}(q||\tilde{q})$ for all defenses with selected hyperparameters (\textbf{A}). Expected coverage for all defenses with selected hyperparameters (\textbf{B}). The performance on well-specified data is shown in black; the colors indicate performance on adversarially perturbed data at certain tolerance levels.}
    \label{fig:appendix_defenses_adv_training_trades}
\end{figure}

% \begin{figure}
%     \centering
%     \includegraphics[width=\textwidth]{figures/main_plot_coverage_rKL.pdf}
%     \caption{Expected coverage for Adversarial training (top) or TRADES (bottom) as defense.}
%     \label{fig:appendix_defenses_adv_training_trades}
% \end{figure}

\begin{figure}
    \centering
    \includegraphics[width=\textwidth]{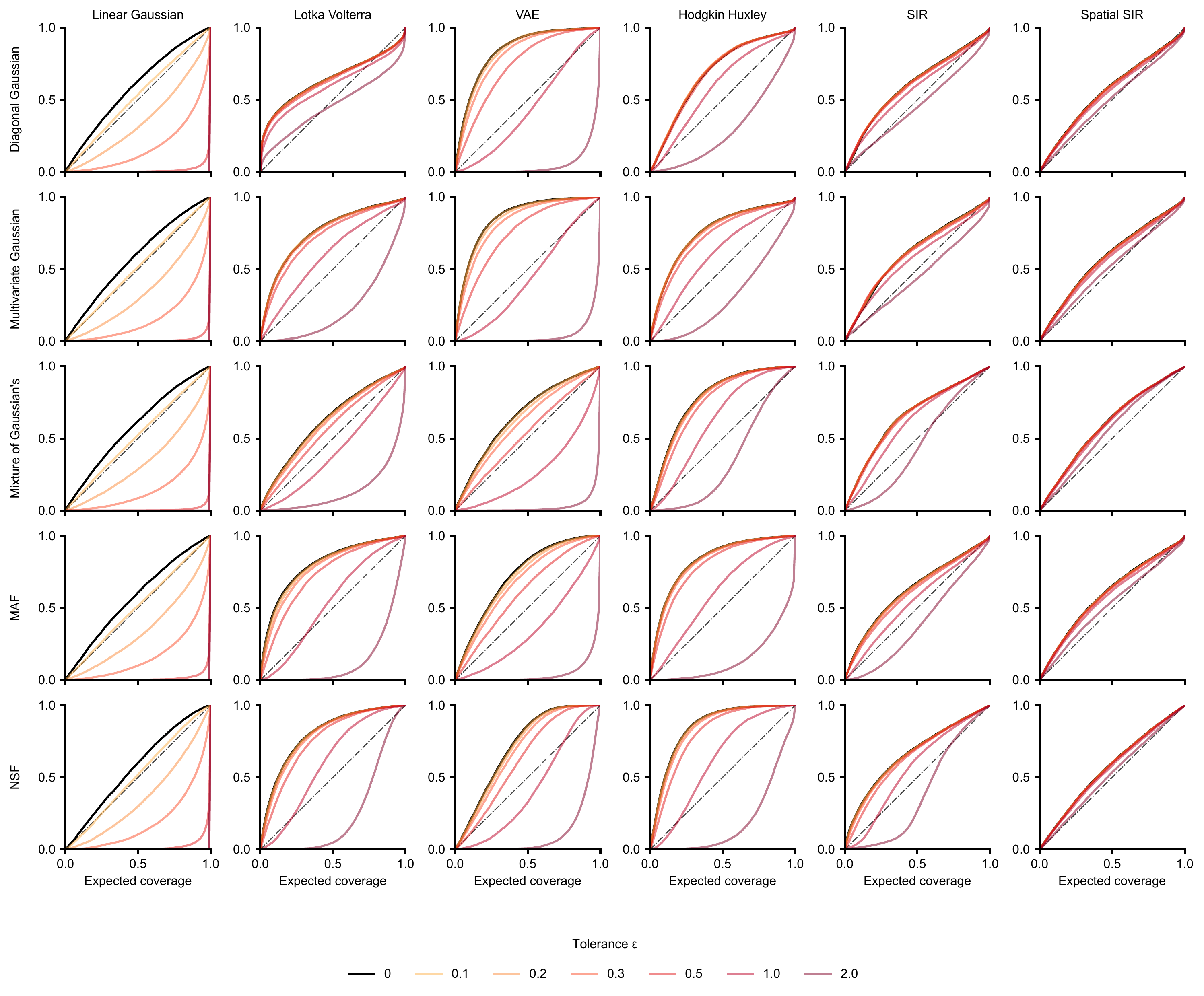}
    \caption{\textbf{Expected coverage for FIM regularization,} Each row shows the expected coverage for a specified density estimator regularized with trace of FIM, at selected hyperparameters. The performance on well-specified data is shown in black; the colors indicate performance on adversarially perturbed data at certain tolerance levels.}
    \label{fig:appendix_defenses_density_estimators}
\end{figure}

\begin{figure}
    \centering
    \includegraphics[width=.8\textwidth]{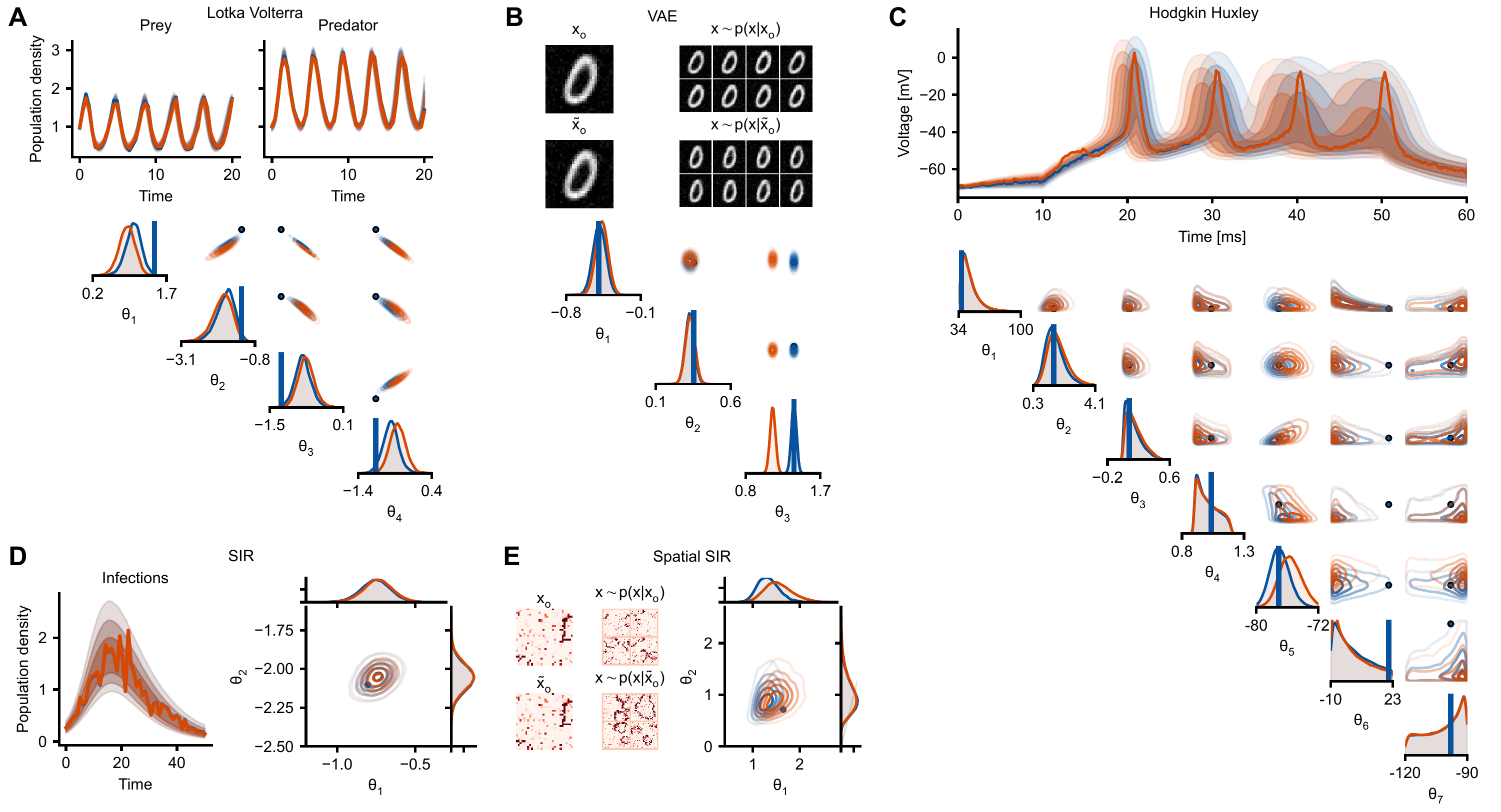}
    \caption{\textbf{Adversarial examples for each benchmark task when employing FIM regularization}. Each panel shows i) the original observation (blue line) and corresponding posterior predictive samples (blue shaded) ii) the adversarial example (orange line) and posterior predictive samples based on the perturbed posterior estimate, and iii) posterior distribution plots with the posterior estimate for the original (blue) and perturbed (orange) data, and the ground-truth parameters (black dot).}
    \label{fig:fig2_appendix_fisher}
\end{figure}

\begin{figure}
    \centering
    \includegraphics[width=.8\textwidth]{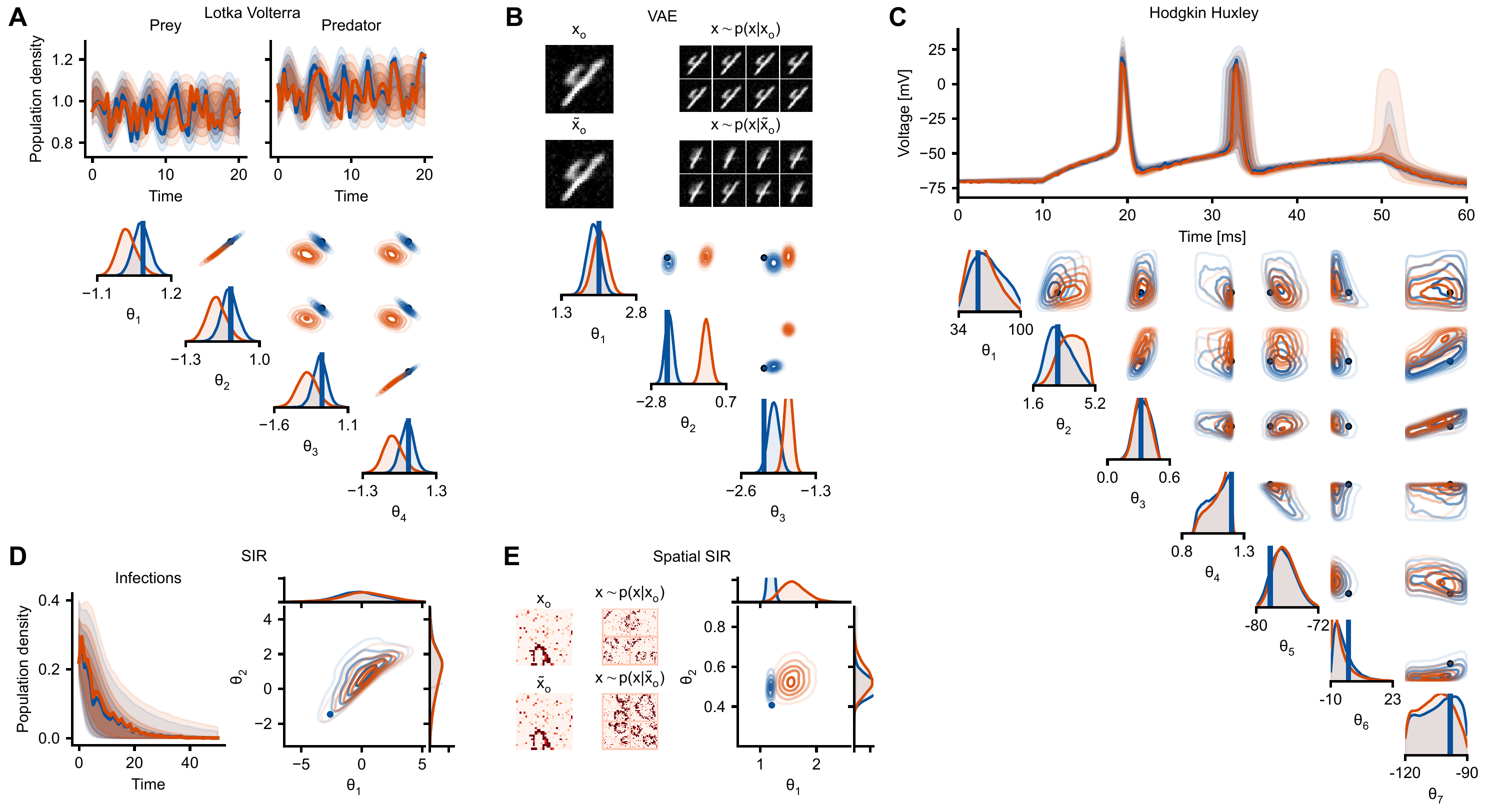}
    \caption{\textbf{Adversarial examples for each benchmark task when using adversarial training}. Each panel shows i) the original observation (blue line) and corresponding posterior predictive samples (blue shaded) ii) the adversarial example (orange line) and posterior predictive samples based on the perturbed posterior estimate, and iii) posterior distribution plots with the posterior estimate for the original (blue) and perturbed (orange) data, and the ground-truth parameters (black dot).}
    \label{fig:fig2_appendix_advtrain}
\end{figure}

\begin{figure}
    \centering
    \includegraphics[width=.8\textwidth]{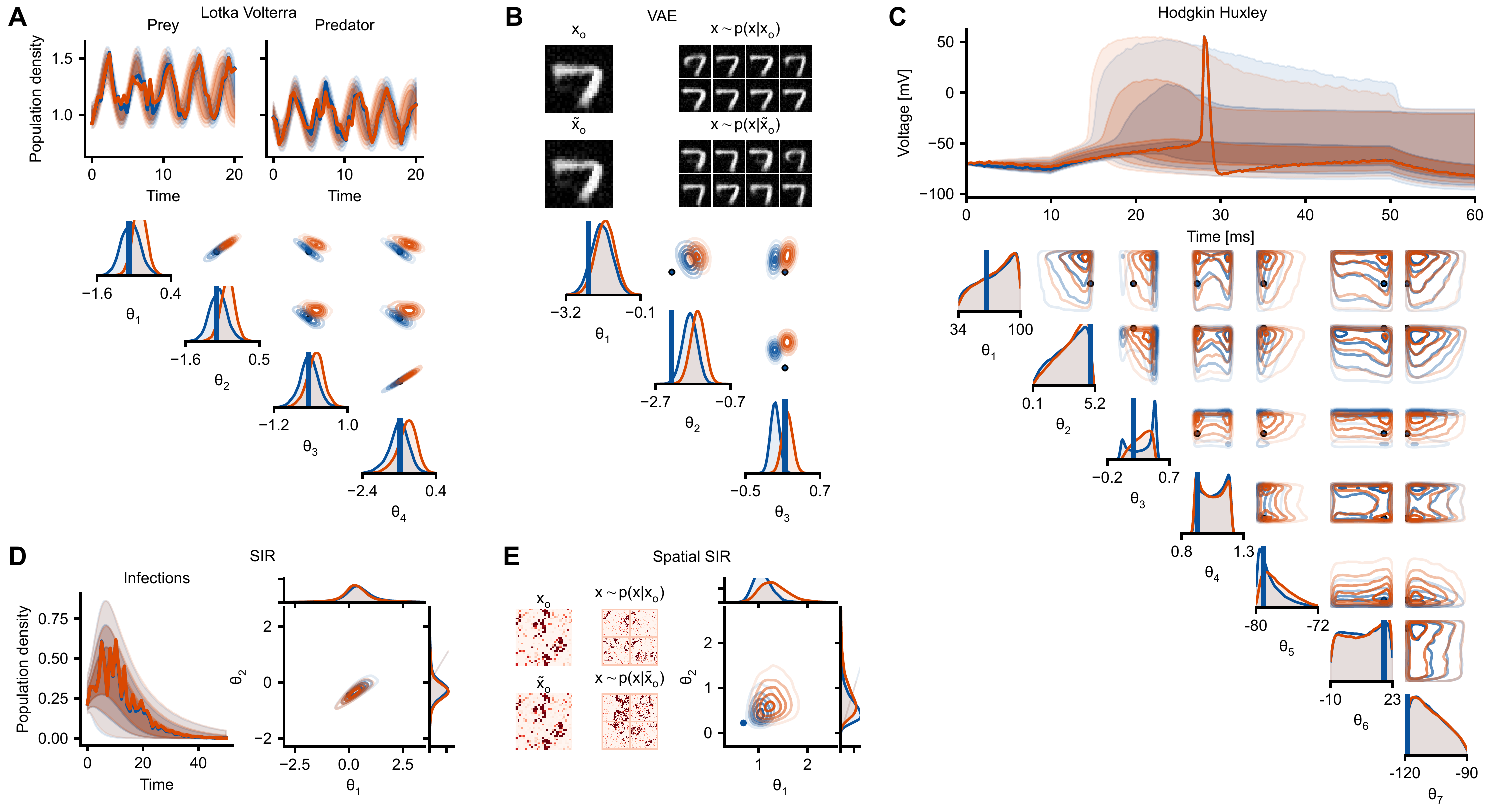}
    \caption{\textbf{Adversarial examples for each benchmark task when using TRADES}. Each panel shows i) the original observation (blue line) and corresponding posterior predictive samples (blue shaded) ii) the adversarial example (orange line) and posterior predictive samples based on the perturbed posterior estimate, and iii) posterior distribution plots with the posterior estimate for the original (blue) and perturbed (orange) data, and the ground-truth parameters (black dot).}
    \label{fig:fig2_appendix_trades}
\end{figure}

\newpage
\section{Optimal attacks for a linear Gaussian simulator}
\label{appendix:gaussian_linear}

The only task in the benchmark with tractable ground-truth posterior for arbitrary $\vx$ is the Gaussian linear task. Here we analyze this task in more detail. We will show that an attack that maximizes the $D_{KL}$ between clean and perturbed posterior corresponds to the strongest eigenvector of the FIM. We then will compare the analytic robustness of the ground truth posterior to the empirical attack on NPE. 

Assume the generative model is given by
$$ p(\vtheta) = \mathcal{N}(\vtheta;\vmu_0, \vSigma_0) \text{ and } p(\vx|\vtheta) = \mathcal{N}(\vx;\bm{A}\vtheta + \bm{b}, \bm{\Lambda}).$$
Then the posterior distribution is well-known and given by
$$ p(\vtheta| \vx_o) = \mathcal{N}(\vtheta; \vmu_p,\vSigma_p)$$
\begin{align*}
    \vmu_p(\vx_o) & =  \vmu_0 + \vSigma_0 \bm{A}^T (\bm{A}\vSigma_0 \bm{A}^T + \bm{\Lambda})^{-1}(\vx_o - (\bm{A}\vmu_0 + \bm{b})) = (\vSigma_0^{-1} + \bm{A}^T\bm{\Lambda}^{-1} \bm{A})^{-1}(\bm{A}^T \bm{\Lambda}^{-1}(\vx_o - \bm{b}) + \vSigma_0^{-1}\vmu_0)\\
    \vSigma_p &= \vSigma_0  - \vSigma_0 \bm{A}^T (  \bm{A}\vSigma_0 \bm{A}^T + \bm{\Lambda})^{-1} \bm{A} \vSigma_0 = (\vSigma_0^{-1}  +  \bm{A}^T \bm{\Lambda}^{-1}  \bm{A} )^{-1} 
\end{align*}

As we will show below, for this simulator, the perturbation which maximizes the $D_{KL}$ between clean and perturbed posterior exactly corresponds to the eigenvector of the Fisher-information matrix with the largest eigenvalue.

\paragraph{Analytical expression for the Kullback-Leibler divergence}

In this model, the Kullback-Leibler divergence $ D_{KL} (p(\vtheta| \vx_o) || p(\vtheta |\vx_o + \vdelta)) $ between clean and perturbed posterior can be computed analytically.

We defined an adversarial example as a distorted observation
given by $\tilde{\vx}_o = \vx_o + \vdelta$. This will only affect the posterior mean, as the covariance matrix is independent of $\vx_o$. The KL divergence between clean and perturbed posterior can be written as
\begin{align*}
D_{KL} (p(\vtheta| \vx_o) || p(\vtheta |\vx_o + \vdelta)) & = 0.5 \cdot \left( tr(\vSigma_{p}^{-1}\vSigma_{p})  - d + \log \left( \frac{|\vSigma_p|}{|\vSigma_p|} \right)\right) \\& \quad + 0.5 (\vmu_p(\vx_o) - \vmu_p(\vx_o + \vdelta))^T \vSigma_p^{-1} (\vmu_p(\vx_o) - \vmu_p(\vx_o + \vdelta))\\
&= 0.5\left( (\vmu_p(\vx_o) - \vmu_p(\vx_o + \vdelta))^T \vSigma_p^{-1} (\vmu_p(\vx_o) - \vmu_p(\vx_o + \vdelta)) \right)\\
\end{align*}
%Note the KL divergence is symmetric in this case. 
and the difference between means can be written as
$$ (\mu_p(\vx_o) - \mu_p(\vx_o + \vdelta)) = (\vSigma_0^{-1} + \bm{A}^T\bm{\Lambda}^{-1} \bm{A})^{-1}\bm{A}^T \bm{\Lambda}^{-1} \vdelta = \vSigma_p \bm{A}^T \bm{\Lambda}^{-1}\vdelta. $$
Hence, we obtain
$$ D_{KL} (p(\vtheta| \vx_o) || p(\vtheta |\vx_o + \vdelta))  = 0.5 \cdot \vdelta^T  \bm{\Lambda}^{-1} \bm{A} \vSigma_p \vSigma_p^{-1} \vSigma_p \bm{A}^T \bm{\Lambda}^{-1}\vdelta= 0.5 \cdot \vdelta^T \bm{\Lambda}^{-1} \bm{A} \vSigma_p \bm{A}^T \bm{\Lambda}^{-1}\vdelta.$$

\paragraph{Analytical expression for the FIM}

Next, we derive a closed-form expression for the Fisher Information Matrix (FIM):

$$ \mathcal{I}_\vx = \mathbb{E}_{p(\vtheta|\vx)} \left[ \nabla_\vx \log p(\vtheta|\vx)(\nabla_\vx \log p(\vtheta|\vx))^T \right]$$

We can write:
\begin{align*}
     \nabla_\vx \log p(\vtheta|\vx) &= -0.5 \nabla_\vx (\mu_p(\vx) - \vtheta)^T \vSigma_p^{-1} (\mu_p(\vx) - \vtheta)\\
     &=- (\nabla_\vx \mu_p(\vx))^T\vSigma_p^{-1}(\mu_p(\vx) - \vtheta) \\
     &=(\vSigma_p \bm{A}^T \bm{\Lambda}^{-1})^T \vSigma_p^{-1}(\vtheta - \mu_p(\vx))\\
     &= \bm{\Lambda}^{1-}\bm{A}(\vtheta - \mu_p(\vx))\\
     &=\bm{\Lambda}^{-1}\bm{A} (\vtheta - \mu_p(\vx))
\end{align*}
Hence the Fisher information matrix with respect to $x$ is given by
\begin{align*}
\mathcal{I}_\vx &=\mathbb{E}_{p(\vtheta|\vx)} \left[ \nabla_\vx \log p(\vtheta|\vx)(\nabla_\vx \log p(\vtheta|\vx))^T \right]\\
&=\mathbb{E}_{p(\vtheta|\vx)} [ \bm{\Lambda}^{-1} \bm{A} (\vtheta - \mu_p(\vx))(\vtheta - \mu_p(\vx))^T \bm{A}^T \Lambda^{-1}]\\
&= \bm{\Lambda}^{-1} \bm{A}\mathbb{E}_{p(\vtheta|\vx)} [ (\vtheta - \mu_p(\vx))(\vtheta - \mu_p(\vx))^T] \bm{A}^T \bm{\Lambda}^{-1}\\
&= \bm{\Lambda}^{-1}\bm{A}\vSigma_p \bm{A}^T \bm{\Lambda}^{-1}
\end{align*}
and thus equivalently
$$ D_{KL} (p(\vtheta| \vx_o) || p(\vtheta |\vx_o + \vdelta))  = 0.5 \cdot \vdelta^T \mathcal{I}_\vx \vdelta$$

This demonstrates that the Kullback-Leibler divergence between clean and perturbed posterior directly corresponds to the Fisher information matrix in the linear Gaussian simulator.

\paragraph{Optimal attack on the linear Gaussian simulator}

When maximizing the $D_{KL}(p(\vtheta|\vx_o) || p(\vtheta |\vx_o + \vdelta))$, the adversary, thus, tries to solve the following problem
$$ \vdelta^* = \max_{\vdelta: ||\vdelta||_2 \leq \epsilon} 0.5\vdelta^T \mathcal{I}_\vx \vdelta. $$
By Reyleight's theorem, this is solved  by $\vdelta^* = \epsilon \bm{v}_{max}$ where $\bm{v}_{max}$ is the eigenvector with the largest eigenvalue of $\mathcal{I}_\vx$ and thus
$$  D_{KL} (p(\vtheta| \vx_o) || p(\vtheta |\vx_o + \vdelta)) \leq 0.5 \lambda_{max} \epsilon^2 .$$
Here $\lambda_{max}$ is the maximum eigenvalue of $\mathcal{I}_\vx$. Thus any empirical attack on the ground truth posterior would be bounded by this quantity.
\begin{figure}
    \centering
    \includegraphics[width=\textwidth]{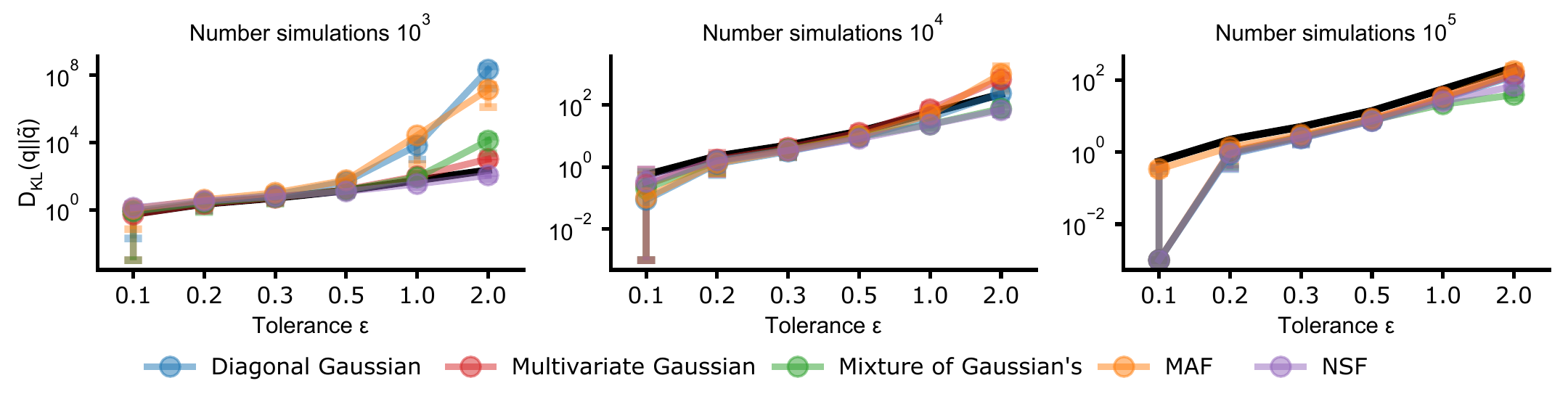}
    \caption{\textbf{Optimal attack on Gaussian linear posterior.} Gaussian linear attacks example on multiple density estimators. We show the "robustness" of the true posterior as a black line.}
    \label{fig:appendix_gl}
\end{figure}

An attack on an inference model that successfully identifies the map to the ground truth posterior should be consistent with this result. Figure~\ref{fig:appendix_gl} illustrates the ground-truth robustness to $\ell_2$ perturbations (black) and the empirical robustness obtained through attacking NPE inference models attempting to solve this task. As the number of simulations increases, the NPE model can more accurately capture the true posterior mapping, resulting in attacks that are bounded by this quantity. This is evident in the figure, where attacks on most models are close to this bound. However, if the model is trained on too few simulations, the results may differ greatly as an incorrect mapping to the posterior is learned, making it more susceptible to adversarial attacks. Interestingly, given insufficient training data, all density estimators tend to be more, and not less, brittle to adversarial perturbations. The attacks are weaker on Mixture models and neural spline flows, which could result from either the attack not being strong enough or the model being too smooth.

\section{Analytical expression for the FIM regularization in generalized linear models}
\label{appendix:defense_analytic}

We analyze the solution on a generalized linear Gaussian density estimator to investigate the bias introduced by FIM regularization. We derive the analytical solutions for the optimal parameters identified by NPE and NPE with FIM regularization and discuss their differences.

Consider a generalized linear Gaussian inference network
$$ q_{\bm{W},\vSigma}(\vtheta|\vx) = \mathcal{N}(\vtheta; \bm{W}\phi(\vx), \vSigma) $$
here $\phi$ is a, possibly nonlinear, \textit{feature mapping} $\phi:\mathbb{R}^{d_\vx} \rightarrow \mathbb{R}^{d_\phi}$. The only learnable parameter is the weight matrix $W \in \mathbb{R}^{d_\vtheta \times d_\phi}$ and covariance matrix $\Sigma \in \mathbb{R}^{d_\vtheta \times d_\vtheta}$.

In this case, the NPE loss using $N$ simulations $(\vx_i, \vtheta_i)$, can be written as
$$ \mathcal{L}(\bm{X}, \bm{\Theta}, \bm{W}, \vSigma) = \frac{1}{2} tr \left(\left( \bm{W} \phi(\bm{X}) -\bm{\Theta}) \right)^T \vSigma^{-1} \left( \bm{W} \phi(\bm{X})-\bm{\Theta} \right)\right)  + \frac{N}{2} \log \det(\vSigma)$$
here $\bm{X} \in \mathbb{R}^{d_x \times N}$ denotes all data points represented as columns of a matrix (equivalently $\bm{\Theta}$). 

Below, we compute analytical expressions for the optimal parameters $\bm{W}, \vSigma$ for (1) NPE and (2) for NPE with FIM-regularization. This allows us to quantify the bias introduced by FIM-regularization in a Gaussian GLM.

\paragraph{Convergence of NPE}
For NPE (without regularization), we can compute the optimal parameters in closed-form:

\begin{align*} \nabla_{\bm{W}} \mathcal{L} &= \vSigma^{-1}(\bm{W} \phi(\bm{X})^T - \bm{\Theta}) \phi(\bm{X})^T \overset{!}{=} 0 \\ \iff& \hat{\bm{W}} = \bm{\Theta} \phi(\bm{X})^T \left( \phi(\bm{X}) \phi(\bm{X})^T \right)^{-1}
\end{align*}
Which is a generalized linear least square regression estimator. Equivalently we can obtain an estimator for the covariance matrix:
\begin{align*} \nabla_\vSigma \mathcal{L} =& -\frac{1}{2} \vSigma^{-1} (\bm{W}\phi(\bm{X}) -\bm{\Theta})(\bm{W}\phi(\bm{X}) -\bm{\Theta})^T \vSigma^{-1} + \frac{N}{2} \vSigma^{-1} \overset{!}{=} 0\\
\iff & N \vSigma^{-1} = \vSigma^{-1} (\bm{W}\phi(\bm{X}) - \bm{\Theta})(\bm{W}\phi(\bm{X}) -\bm{\Theta})^T \vSigma^{-1}\\
\iff & \hat{\vSigma} = \frac{1}{N} (\bm{W} \phi(\bm{X}) - \bm{\Theta}) (\bm{W}\phi(\bm{X}) -\bm{\Theta})^T.
\end{align*}
As $\hat{\bm{W}}$ is estimated independently of $\vSigma$, we can plug in $\bm{W} = \hat{\bm{W}}$ to globally minimize the loss.

\paragraph{Convergence of NPE with FIM regularization}
In this model, we can also compute the FIM regularized solution in closed-form. The FIM is given by

$$ \mathcal{I}_\vx = J_\phi(\vx)^T \bm{W}^T \vSigma^{-1} \bm{W} J_\phi(\vx).$$
 Here $J_\phi(\vx)$ denotes the Jacobian matrix of $\phi$ at $\vx$. Hence the FIM regularized model minimizes the loss
$$ \mathcal{L}_{FIM}( \bm{X},\bm{\Theta}, \bm{W},\vSigma, \beta) = \mathcal{L}(\bm{X},\bm{\Theta}, \bm{W}, \vSigma) + \frac{\beta}{N} \sum_{i=1}^N  tr(\mathcal{I}_{\vx_i}) .$$
To avoid clutter in notation, let $\Omega(\bm{X}) = \frac{1}{N} \sum_{i=1}^n J_\phi(\vx_i) J_\phi(\vx_i)^T$. Then we can write a solution that minimizes this loss as 

\begin{align*} \nabla_{\bm{W}} \mathcal{L}_{FIM} =& \nabla_{\bm{W}}  \mathcal{L}  + 2 \beta \vSigma^{-1} \bm{W}\Omega(\bm{X})\overset{!}{=} 0 \\ \iff &  \hat{\bm{W}}_{FIM} = \bm{\Theta} \phi(\bm{X})^T \left( \phi(\bm{X})\phi(\bm{X})^T + 2 \beta \Omega(\bm{X})\right)^{-1} 
\end{align*}

and

\begin{align*}
    \nabla_\vSigma \mathcal{L}_{FIM} = & \nabla_\vSigma \mathcal{L} - \beta \vSigma^{-1} \bm{W} \Omega(\bm{X}) \bm{W}^T \vSigma^{-1} \overset{!}{=} 0 \\
    \iff & \hat{\vSigma}_{FIM} = \hat{\vSigma}+ \frac{4\beta}{N} \bm{W}\Omega(\bm{X}) \bm{W}^T.
\end{align*}

Again we can plug in $\hat{\bm{W}}_{FIM}$ to globally minimize the loss.

\paragraph{Bias introduced by FIM-regularization}

The solution for $W$ corresponds to a Tikhonov regularized least squares solution, which simplifies to ridge regression in the linear case \citep{golub1999tikhonov}.  Recent research has shown a connection between adversarial training and ridge regression in the linear-least squares setting. \citep{ribeiro2022surprises}. The regularization approach based on Fisher Information Matrix (FIM) incorporates a bias term through the average Jacobian outer product, which is known to recover those directions that are most relevant to predicting the output \citep{trivedi2020expected}. Thus, the regularization strength is directed towards the directions to which the feature mapping $\phi$ is most sensitive. This bias increases monotonically with the regularization parameter, leading to an asymptotically to smooth mean function.

Additionally, FIM regularization overestimates the covariance matrix for a finite number of data points. Interestingly, as $N \rightarrow \infty$ the additive bias on the covariance matrix vanishes. 
%This makes intuitive sense because the parameterization of the density estimator makes $\vSigma$ independent of $\vx$ and hence is not affected by perturbations of $\vx$ if the posterior mean is correctly estimated. 
This is in agreement with our empirical results of `conservative' posterior approximations.

\section{FIM approximations}
\label{sec:FIM}

We propose an efficient approximation method to scale FIM-based approaches to complex density estimators. Here we investigate the effect of the approximation compared to an exact method. In order to compare our method to a closed-from FIM, we use a simple Gaussian density estimator of the form
$$ q_\phi(\vtheta|\vx) = \mathcal{N}(\vtheta; \vmu_\phi(\vx), diag(\bm{\sigma}^2_\phi(\vx))). $$
Here $\vmu: \mathbb{R}^{d_\vx} \rightarrow \mathbb{R}^{d_\vx}$ is the mean function parameterized as neural network and $\bm{\sigma}: \mathbb{R}^{d_\vx} \rightarrow \mathbb{R}^{d_\vx} $ the standard deviation, which is transformed to a diagonal covariance matrix.
Let us denote the distributional parameters as
$$ \veta = \binom{\vmu_\phi(\vx)}{\bm{\sigma}_\phi(\vx)}$$
and let $J$ be the Jacobian matrix defined by $J_{ij} = \frac{\partial \veta_i}{ \partial x_i}$. Then we can write $ \mathcal{I}_\vx = J^T \mathcal{I}_\veta J$. Notice that $ \mathcal{I}_\veta$ is a Gaussian FIM with respect to the parameter $\vmu$ and $\bm{\sigma}$ which is given by
$$ \mathcal{I}_\veta = -\mathbb{E}_{q_\veta} \left[ \nabla^2_\veta \log q_\veta(\vtheta) \right ] = \begin{pmatrix}
    diag(\frac{1}{\bm{\sigma}^2}) & 0 \\ 
    0 & diag(\frac{2}{\bm{\sigma}^2})
\end{pmatrix} . $$
The Jacobian matrix can be computed via autograd and thus $\mathcal{I}_\vx$ can be computed for any given $\vx$.

To test how well our approximations work, we tested the following regularizers:
\begin{itemize}
    \item FIM Largest Eigenvalue: Using the regularize $\Omega(\vx) = \beta \cdot \lambda_{\max}(\mathcal{I}_\vx)$ with the exact FIM. 
    \item FIM Trace: This uses $\Omega(\vx) = \beta \cdot tr(\mathcal{I}_\vx)$ also with the exact FIM.
    \item FIM Trace [EMA]: This uses  $\Omega(\vx) = \beta \cdot tr(\hat{\mathcal{I}}_\vx)$ estimated as described in the main paper.
\end{itemize}

\begin{figure}
    \centering
    \includegraphics[width=0.8\textwidth]{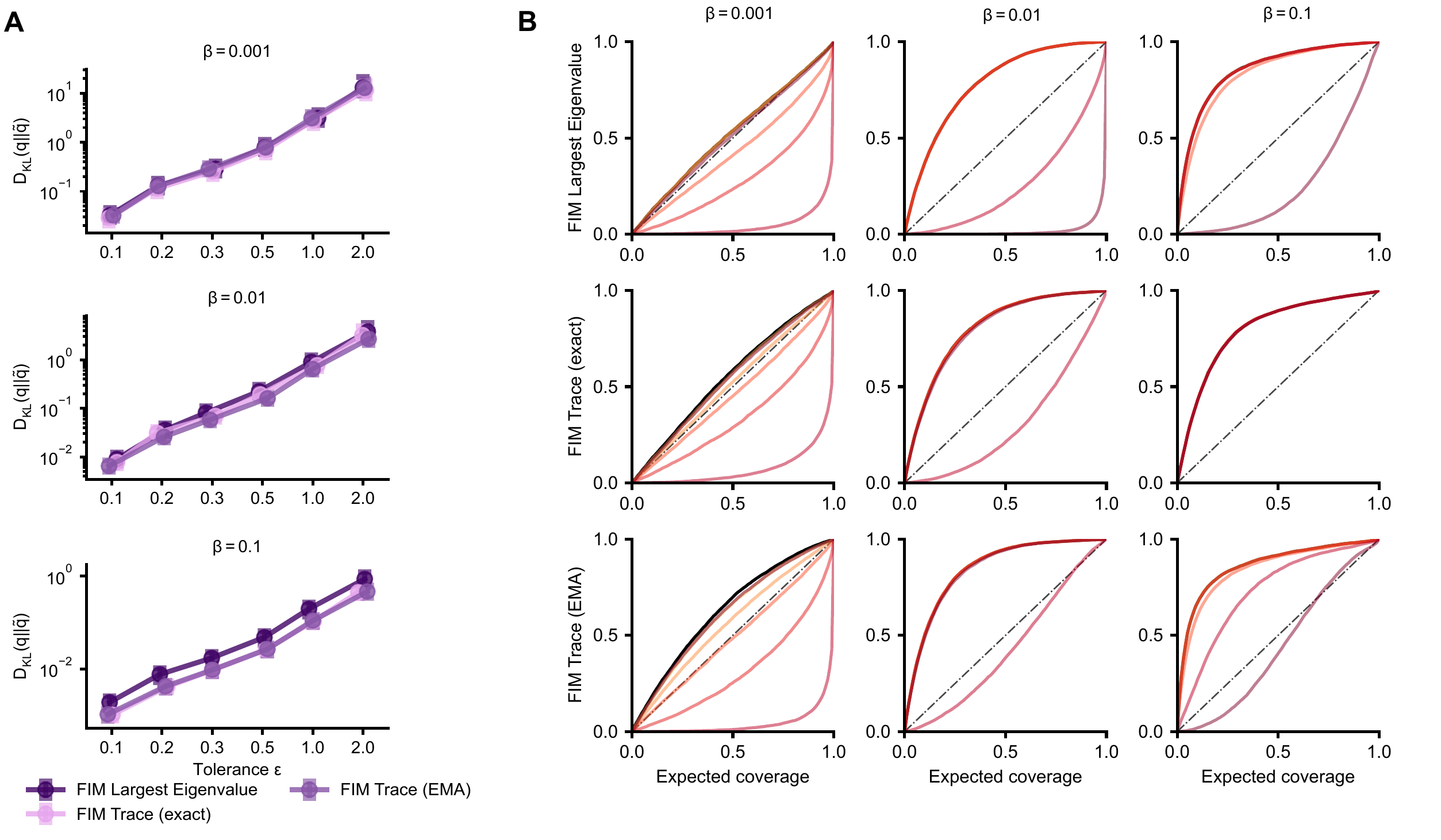}
    \caption{\textbf{FIM approximation:} Columns show the robustness and expected coverage for different choices of regularization strength $\beta$. Top row shows robustness $D_{KL}(q||\tilde{q})$. The bottom rows show the expected coverages of the FIM Largest Eigenvalue, FIM Trace (exact), and FIM Trace (EMA) regularizers.}
    \label{fig:mc_approx}
\end{figure}

The results on the VAE task are shown in Figure \ref{fig:mc_approx} using three different choices $\beta = 0.001, 0.01, 0.1$. Notably, for small regularization strengths, all techniques work similarly. Both trace-based regularizers decrease $D_{KL}(q||\tilde{q})$ more strongly, as expected by the fact that they are upper bounds of the largest eigenvalue. Notably, train time increased to five hours using the exact eigenvalue, compared to under two minutes using our approach. 

Overall, these results demonstrate that our approximations to the largest eigenvalue of the FIM incur only a small cost in adversarial robustness.

\section{Comparission to MCMC based posteriors}
\label{sec:appendix_true_post}

We emphasize metrics that can be efficiently computed without requiring direct access to the true posterior. This choice is justified because numerous tasks lack a tractable or "expensive to evaluate" likelihood. Consequently, calculating the posterior individually for thousands of observations would be computationally expensive. However, it is worth noting that we have a likelihood for some of the benchmark tasks and can perform non-amortized posterior computation with a manageable computational burden.

For the Gaussian Linear task, we have the advantage of an analytic solution. This simplifies the computation of the posterior, as we can directly derive the necessary quantities without relying on approximation methods or sampling techniques. In the case of the SIR, VAE, and Lotka Volterra tasks, we have tractable likelihood functions. This enables us to estimate the posterior distribution using likelihood-based inference methods, such as MCMC. We use a two-stage procedure to get good posterior approximations. First, we run one hundred parallel MCMC chains initialized from the prior. For the SIR and Lotka Volterra, we used an adaptive Gaussian Metropolis Hasting MCMC method \citep{andrieu2008tutorial}. For the VAE task, we used a Slice sampler \citep{neal2003slice}. In the second stage, these results were used to train an unconditional flow-based density estimator, which then performs Independent Metropolis Hasting MCMC to obtain the final samples \citep{holden2009adaptive}.

\begin{figure}
    \centering
    \includegraphics[width=\textwidth]{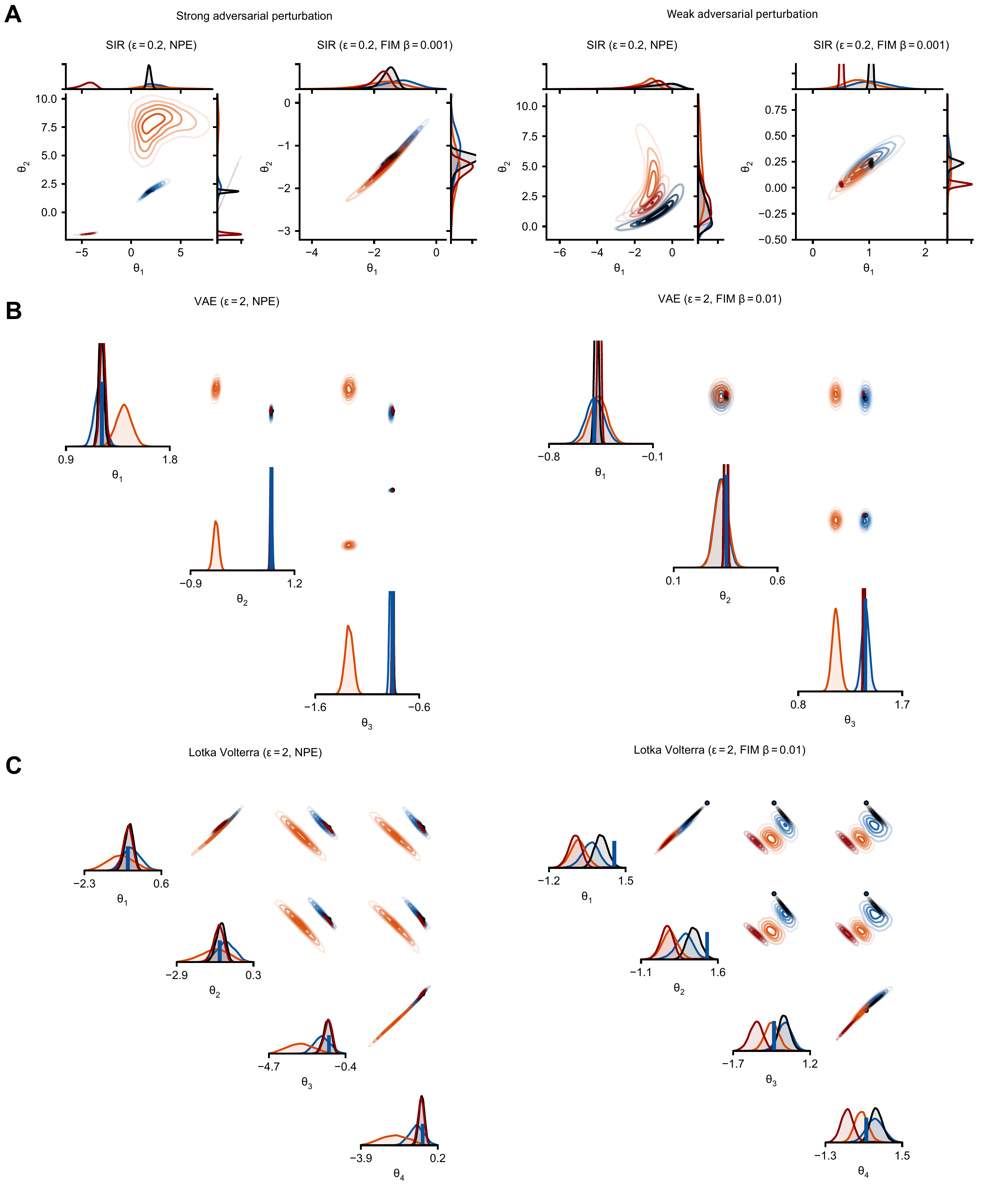}
    \caption{\textbf{Approximate posterior distributions plotted against MCMC-based posterior estimate.} This figure compares approximate amortized posterior estimates and the posterior obtained through MCMC sampling. In blue and orange, we can see the approximate amortized posterior estimate on the clean observations $q_\phi(\vtheta|\vx)$ and a perturbed observation $q_\phi(\vtheta|\tilde{\vx})$. In black, we plot the true posterior $p(\vtheta|\vx)$ and in red, the true adversarial posterior $ p(\vtheta|\tilde{x})$ (both estimated via MCMC). We present this comparison for the SIR task, showcasing both a strong adversarial perturbation and weaker perturbation (as measured by $D_{KL}(q||\tilde{q})$). For the VAE and Lotka Volterra tasks, we only display the results for strong perturbations (denoted as \textbf{B} and \textbf{C}, respectively). } 
    \label{fig:fig_true_posterior}
\end{figure}

We present an illustration of several approximated  (adversarial)  posteriors alongside their corresponding ground truth obtained through Markov Chain Monte Carlo (MCMC) methods (Figure \ref{fig:fig_true_posterior}). As discussed in the main paper, even minor perturbations can lead to significant misspecification of the SIR model. We observe substantial changes in the true posterior distribution due to adversarial perturbations (\ref{fig:fig_true_posterior}A). Notably, the standard NPE method appears to be highly susceptible to this scenario, often producing predictions that seem arbitrary. In contrast, the FIM regularized inference networks exhibit a similar trend as the true posterior; however, they tend to be underconfident in their predictions as expected. 

Figure \ref{fig:fig_true_posterior}B illustrates the behavior of the variational autoencoder (VAE) task in the presence of adversarial perturbations. Interestingly, the ground-truth posterior distribution remains remarkably unaffected by these perturbations, exhibiting a high degree of invariance. Yet, the inference network, responsible for approximating the posterior, proves to be susceptible to being deceived by these adversarial perturbations. This effect is strongly reduced for FIM regularization. We observe this similarly on the Lotka Volterra task.

\begin{figure}
    \centering
    \includegraphics[width=\textwidth]{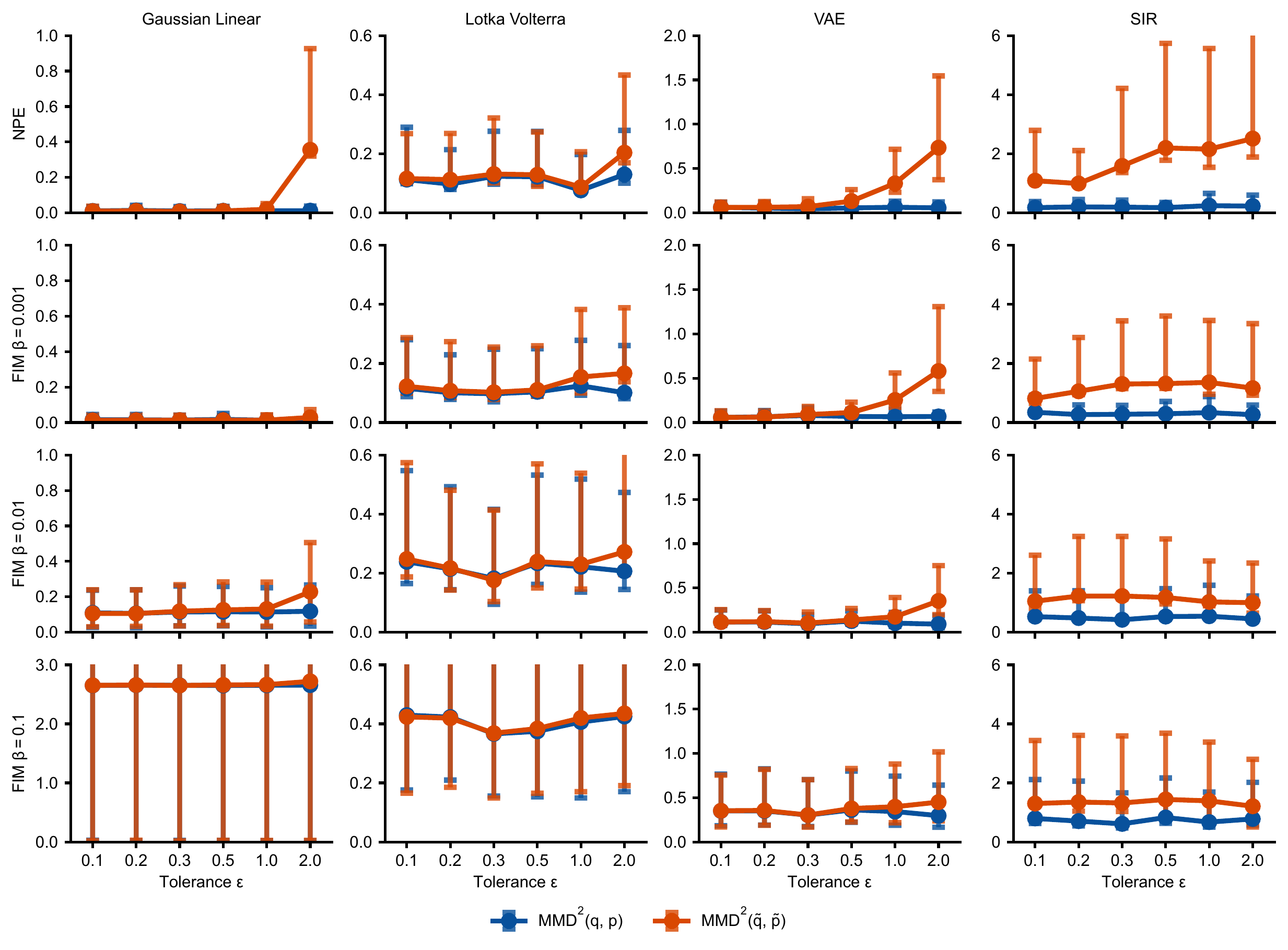}
    \caption{\textbf{MMD distance against MCMC-based posteriors on clean and perturbed data.}. This figure showcases the squared MMD distance between the estimated posterior and the posterior obtained through MCMC. The blue line represents the MMD distance on well-specified data, while the orange line illustrates the MMD distance on adversarially perturbed observations. These metrics were computed from a randomly selected subset of 100 clean and perturbed observations.}
    \label{fig:metric_true_posterior}
\end{figure}

We quantified this difference on a randomly selected set of 100 pairs $(\vx, \tilde{\vx})$ by computing $ MMD^2(q, p) = MMD^2 (q_\phi(\vtheta|\vx), p_{MCMC}(\vtheta|\vx))$ and $ MMD^2(q, p) = MMD^2 (q_\phi(\vtheta|\tilde{\vx}), p_{MCMC}(\vtheta|\tilde{\vx}))$ (Figure~\ref{fig:metric_true_posterior}). Consistent with expectations, the MMD value for well-specified data demonstrates a relatively good agreement between the true posterior and the inference network's approximation across different tasks. However, when confronted with adversarially perturbed data, the MMD value increases significantly. This indicates that the inference network struggles to accurately capture the underlying posterior distribution in the presence of adversarial perturbations. These findings highlight the limitations of the inference network and its susceptibility to such adversarial perturbations.  FIM regularization can mitigate this effect, effectively reducing the impact of adversarial perturbations on the inference network's performance. However, this improvement comes at the expense of decreased approximation quality on well-specified data.

\end{document}